# A Perspective Analysis of Handwritten Signature Technology

Moises Diaz, Miguel A. Ferrer, Donato Impedovo, Muhammad Imran Malik, Giuseppe Pirlo, Réjean Plamondon.

**Abstract**

Handwritten signatures are biometric traits at the center of debate in the scientific community. Over the last 40 years, the interest in signature studies has grown steadily, having as its main reference the application of automatic signature verification, as previously published reviews in 1989, 2000, and 2008 bear witness. Ever since, and over the last 10 years, the application of handwritten signature technology has strongly evolved, andmuch research has focused on the possibility of applying systems based on handwritten signature analysisand processing to a multitude of new fields. After several years of haphazard growth of this research area, it is time to assess its current developments for their applicability in order to draw a structured way forward. This perspective reports a systematic review of the last 10 years of the literature on handwritten signatures with respect to the new scenario, focusing on the most promising domains of research and trying to elicit possible future research directions in this subject.

## 1. INTRODUCTION

During the last 40 years, a number of comprehensive surveys and state-of-the-art reviews in automatic signature verification (ASV) [170] have been published. Among a large number of papers,a couple of relevant academic articles have had a strong impact. One of the first state-of-the-art papers in this field was published in 1989 by Plamondon and Lorette [214]. Approximately 10 yearslater, Plamondon and Srihari published another survey [218]. In a similar span of time, Impedovo and Pirlo [123] published a comprehensive survey in 2008.

From this perspective, it can be deduced that 10 years is an acceptable period of time to make a substantial upgrade to the automatic signature verification state of the art. With this in mind, we review the technology in this article, taking into account the novel advances and emerging issues of ASV in the last 10 years, from 2008 up to now.

Moreover, shorter state-of-the-art papers have been published, mostly in conference proceedings or book chapters (e.g., [57, 80, 111, 126, 196]), further explaining specific advances in signature verification and also pointing out several research directions and milestones. The scope of thepresent article aims at a wider breadth of consideration as compared with the published reviewsin conferences or book chapters.

In fact, the spreading of low-cost technologies for handwriting acquisition and processing (PDAs, tablets, smartphones, etc.), the excellent results obtained in a multitude of research on handwritten signature analysis and processing, and the high acceptability for daily use of the sig- nature for authentication (many countries have articulated legal procedures for using them) allowsus to look at the handwritten signature as a biometric trait supporting a multitude of applicationsin several domains. Starting from this consideration, the aim of this article is, on the one hand,to introduce a systematic review of the advancement of research in the last decade in the field of handwritten signature analysis and processing, with respect to both traditional applications and new application domains.

On the other hand, the use of the handwritten signature, in many cases worldwide, is mandatory. It has a double legal importance: first, it establishes the identity of a person; second, it states the intention of that person. Sometimes more than a handwritten signature is required, such as a signature in the presence of a public notary or lawyer. So far, its use is well recognized by administrative and financial institutions [258] and it is well accepted [123].

The General Data Protection Regulation (GDPR 2018) establishes the new rules for personal dataprocessing within the European Union as well as for their use outside the EU borders. GDPR2018 isexpected to impact on the biometric research field. In fact, in the GDPR context, biometric data areconsidered a special category of personal data (article 9): processing of biometric data is prohibiteduntil the subject has given explicit consent or it is authorized by law (article 9, par. a & b). Moreover,since the processing of such personal data implies high risks, a data protection impact assessment is required (article 35). A deep insight regarding the implications of the GDPR with respect to research activities can be found in [36], and also with respect to novel applications of signatures [216]. It appears quite clear that the assessment of the consent form for the acquisition phase is a nontrivial problem, as is the one-off acquiring a license agreement for research dataset release.

## 1.1. Classical Scheme of Automatic Signature Verification

A typical ASV scheme is illustrated in Figure 1. When an individual signs into an ASV system, a particular tool (e.g., pen and paper) is used and a sensor $S_i$ typically digitizes the sample. Thereare a large number of possible sensors for digitizing the signature, e.g., a scanner, a digital tablet,a PDA, a mobile phone, and even a specialized pen. In addition to capturing the signature, someperturbations usually affect the acquired sample. The perturbations are external factors that refer to the tool used to sign, the posture of the individual, and the jewelry worn, environmental changes, such as background noise, or sensor limitations, including resolution or sample rate.

Another factor that modifies the signature sample during the repetitions is intrapersonal variability. Intrapersonal variability means the dissimilarity between signatures executed by the same writer. It is an internal factor, which might refer to the emotions, stress, tiredness, alcohol or drug influences, neuromotor conditions, biological aging effects, cognitive-motor impairment, and other factors as a consequence of the individual's mood, the time available to write the signature, or the willingness to cooperate. Therefore, we can describe a measured signature as the sum of external and internal perturbations $\hat{v}_i$ over the execution of the signature action plan learned over the years through practice.

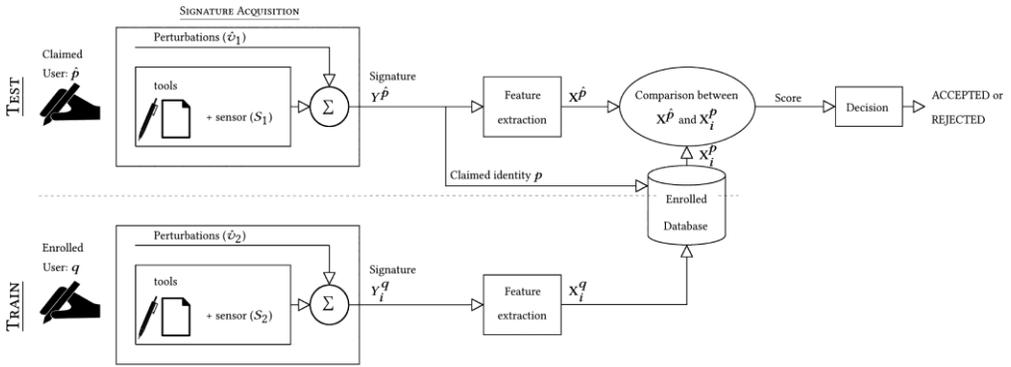

*Fig. 1. Typical scheme of an automatic signature verifier (ASV).*

In an ASV, there are two operating modes: training and testing. In the training mode, a genuine user, let's say q, provides one or several repetitions of his signature, denoted as $Y_i^q$. These signatures are parameterized and their parameters, $X_i^q$, are stored in a database. In the testing mode, the ASV builds a feature matrix ($X^{\hat{p}}$) from questioned user $\hat{p}$'s signature $Y^{\hat{p}}$. Since this questioned user claims to be previously enrolled subject p, a comparison of the signature's parameters $X^{\hat{p}}$ and the signature's parameters $X_i^p$ of enrolled user p is worked out. As a result, a score is obtained. Finally, the ASV makes a decision by accepting or rejecting the claimed hypothesis of $\hat{p}$= p, usually based on a decision threshold. Figure 1 shows these stages using this nomenclature.

Beyond the intrapersonal variability, the greatest challenge faced by ASV systems is the unpredictable interpersonal variability. This means the similarity between signatures executed by different writers. Interpersonal variability can be reduced by faking the identity of signers through three main kinds of forgeries: (1) *Random Forgeries (RF):* This is the situation in which an impostor, without previous knowledge of a specific signature, tries to verify the identity of a signer by using his or her own genuine signature. (2) *Simple Forgery (SF):* This occurs when the forger knows the writer's name but has no access to a sample of the signature. Its importance in signature verifica-tion depends on the country of origin of the signature style. In some countries, many people sign simply by writing their own name and surname, whereas in other countries signers design illegible signatures with complex flourishes or rubrics. (3) *Skilled Forgeries (SK):* These are produced by an impostor who has learned the signature of his or her victim and tries to reproduce it with a similar intraclass variability. The test for this type of forgery is the most relevant in signature verification. Additionally, it is worth mentioning that there are two modalities in signatures: (1) *Offline (or static) signatures:* Here the signature information is typically contained in a scanned image where the inked signature is deposited on a piece of paper through the use of a tool held in the hand like a pen or pencil, among others. (2) *Online (or dynamic) signatures:* Their main characteristic is that they contain the temporal and dynamic order in which the signer executed the signature. To register this kind of signature, a device such as a digitizing tablet is required.

Although all the ASV systems operate according to the general scheme illustrated in Figure 1, the strategy for extracting the features and producing the final classification is generally different between online and offline systems. Nevertheless, to test an ASV, a meaningful statistical signature database is required in both cases.

The article is organized as follows: Section 2 discusses aspects related to handwritten signature databases, which have become a crucial aspect of the research. Best practices in automatic sig- nature verification is discussed in Section 3. In Section 4, a summary of the main competition results is reported by highlighting the state of the art in the field. The forensic aspects related to signature analysis are discussed in Section 5. Some of the most promising new research directions are addressed in Section 6. Section 7 concludes on a positive and optimistic note regarding the future of this technology.

## 2. DATABASES

Handwritten signatures are the product of a complex human process. There are, therefore, innumerable variables and parameters that cannot be represented in a single signature database of a practical size. ASVs typically model the peculiarities in the signatures found in a particular signature database, and this limits their applicability. This effect probably leads to biased results since not all human behavior is taken into account in the datasets.

As such, the challenge is not only to improve the state-of-the-art results with current databases but also to develop incremental signature databases, which consider more and more examples of human behavior: temporal evolution, aging, posture, emotions, multiscripts, collection devices, user skills in using new devices, occlusions (e.g., lines, printed or handwritten text or stamps), and so on. As more aspects and conditions of human behavior are considered, the ASV algorithm is tested more realistically. For a reliable test of current and forthcoming algorithms, large projects to collect huge handwritten signature databases or open government data initiatives are needed.

These real databases must be built according to ethical aspects such as the privacy of the donors. This issue hinders the acquisition of the signatures (the signers should be asked to sign an agreement) and database maintenance (the signers should have the right to remove their signature from the database at their convenience). To alleviate this problem, an alternative has been to develop databases with disguised genuine signatures, where the signers invent a pretended genuine signature (e.g., [140, 271]). The main consideration of these signatures is that the donors have not overlearned the trajectory of their own signatures. Such lack of learning leads to the generation of genuine signatures with unrealistic properties (e.g., poor kinematic properties).

Another interesting solution is the development of synthetic signature databases that allow the integration of a large number of users under a wide variety of situations. In this case, studies of the human neuromotor system are necessary to design a human-like synthesizer. Obviously, synthetic databases can be useful until an equivalent real database is developed.

We present below a list of publicly available databases, give a description of their common limitations, and point out the best practices to use such signature databases in automatic signature verification.

## 2.1. Most Popular Publicly Available Handwritten Signature Databases

There is a long list of signature databases used in research. In this article, we have included most publicly available corpuses since they are more accessible for creating common benchmarks and for comparing results. These signature databases are listed in Table 1. Beyond pointing out some characteristics of each database, we have also included a similar description of some databases often reported in some studies, but which are not freely distributed at this time. A controversial aspect to take into account is the way to label the types of forged signatures in the databases. As previously mentioned, depending on the research project, the forgeries can be referred to as skilled forgeries [16], deliberate forgeries [75], disguise [150] or random or impostor signatures [101], or simulated or highly skilled forgeries [191].

This nomenclature suggests a categorization of the degree of closeness to genuine signatures. It is worth pointing out that the majority of databases have been collected in laboratories. This implies that the forgeries are typically produced by non- professional forgers or by volunteers like students or other staff.

In Table 1, we can observe that the majority of corpuses belong to Western scripts.

*Table 1. Description of the Most Common Handwritten Signature Databases*

| Corpus | Modality | Script | Characteristics | Distinction |
|---|---|---|---|---|
| **Common publicly available offline signature databases** | | | | |
| MCYT-75 [81, 191] | Offline | Western | G:1125, F:1125, W:75 | Taken in two sessions |
| CEDAR [134] | Offline | Western | G:1320, F:1320, W:55 | |
| 4NSigComp2010 [152] (Scenario 1) | Offline | Western | G:113, F:194, D:27 W:2 | 300dpi, BMP |
| 4NSigComp2012 [150] | Offline | Western | G:113, F:273, D:64, W:3 | 300dpi, BMP |
| SigWIcomp2015 [160] | Offline | Western (Italian) | G:479, F:249, W:50 | Taken in several sessions |
| Tunisian SID-Signature [2] | Offline | Tunisian | G:6000, F:4000, W:100 | 300dpi. Collected in different sessions |
| SigWIcomp 2015 [160] | Offline | Bengali | G:240, F:300, W:10 | |
| Off QU-PRIP [6] | Offline | Arabic | G:12000, W:1017 | 600dpp. JPEG. Several nationalities |
| UTSig [241] | Offline | Persian | G:3105, F:4830, D:345, W:115 | |
| PHBC [5] | Offline | Persian | G:1000, F:200, W:100 | 300dpi, tiff |
| BHSig260 [192] | Offline | Bengali | G:2400, F:3000, W:100 | |
| BHSig260 [192] | Offline | Devanagari | G:3800, F:4800, W:160 | |
| **Common publicly available on-line signature databases** | | | | |
| MCYT-100 sub-corpus [81, 191] | Online | Western | G:2500, F:2500, W:100 | This is a subset of online MCYT-330 |
| SUSIG-Visual sub-corpus [140] | Online | Western | G:1880, F:940, W:94 | |
| SUSIG-Blind sub-corpus [140] | Online | Western | G:820, F:880, W:88 | Volunteers did not see visual feedback while signing |
| SG-NOTE database [167] | Online | Western | G:500, W:25 | Taken in two-sessions |
| e-BioSign-DS1-Signature DB [246] | Online | Western | G:520 × 5, F:390 × 5, W:65 × 5 | 5 different writing tools (stylus and finger). Taken in two sessions |
| BSEC 2009 [119] | Online | Western | G:6480, F:4320, W:432 | Taken in two sessions. Digitizing tablet and a PDA |
| SigWIcomp 2015 [160] | Online | Western (German) | G:450, F:300, W:30 | |
| MOBISIG database [14] | Online | Western (Hungarian) | G:3735, F:1660, W:83 | Finger-drawn signatures. Donors did not use their real signature |
| Signature Long-Term DB [95] | Online | Western | G:1334, W:29 | Acquisition in a 15-month time span |
| SVC-Task1 subcorpus [271] | Online | Chinese and English | G:800, F:800, W:40 | Only dynamic trajectory. Donors did not use their real signature |
| SVC-Task2 subcorpus [271] | Online | Chinese and English | G:800, F:800, W:40 | Pressure and pen inclination. Donors did not use their real signature |
| NDSD [267] | Online | Persian | G:3575, F:2200, W:55 | Taken in two sessions, professional forgers |
| **Common publicly available dual online and offline signature databases** | | | | |
| BiosecurID-SONOF [91] | On- and offline | Western | G:2112 × 3, F:1584 × 3, W:132 × 3 | Taken in four sessions. Real on/off and synthetic off data |
| NISDCC database [8, 28] | On- and offline | Western | G:1200 × 2, F:600 × 2, W:100 × 2 | Real on/off data |
| Synthetic MCYT-300 [75] | On- and offline | Western | G:8250 × 2, F:8250 × 2, W:330 × 2 | Imitation of the performance and appearance of [191] |
| Synthetic BiosecureID-UAM [75] | On- and offline | Western | G:2112 × 2, F:1584 × 2, W:132 × 2 | Imitation of the performance and appearance of [91] |

(Continued)

*Table 1. Continued*

| Synthetic NISDCC [75] | On- and offline | Western | G:1200 × 2, F:600 × 2, W:100 × 2 | Imitation of the performance and appearance of [8, 28] |
|---|---|---|---|---|
| Synthetic SVC2004 [75] | On- and offline | Western | G:1600 × 2, F:1600 × 2, W:80 × 2 | Imitation of the performance and appearance of [271] |
| Synthetic SUSIG-Visual [75] | On- and offline | Western | G:1880 × 2, F:940 × 2, W:94 × 2 | Imitation of the performance and appearance of [140] |
| Synthetic SUSIG-Blind [75] | On- and offline | Western | G:820 × 2, F:880 × 2, W:88 × 2 | Imitation of the performance and appearance of [140] |
| SigComp2009 [28] | On- and offline | Western | G:1260 × 2, F:2613 × 2, W:112 × 2 | NISDCC and NFI datasets |
| SigComp2011 [151] | On- and offline | Western (Dutch) | G:3162 × 2, F:1491 × 2, W:128 × 2 | 400 dpi, PNG |
| SigWiComp2013 [161] | On- and offline | Western (Dutch) | G:270 × 2, F:972 × 2, W:27 × 2 | |
| SigComp2011 [151] | On- and offline | Chinese | G:946, F:1598, W:40 | 400 dpi, PNG |
| SigWiComp2013 [161] | On- and offline | Japanese | G:2604 × 2, F:2232 × 2, W:62 × 2 | |
| Off/On QU-PRIP [6] | On- and offline | Arabic | G:1164 × 2, D:582 × 2, F:200 × 2, W:194 × 2 | Mostly Arabic donors |
| SIGMA. Malaysian [3] | On- and offline | Malaysian | G:6000 × 2, F:2000 × 2, W:265 × 2 | Taken in three sessions |
| Synthetic OnOff SigBengali-75 [47, 74] | On- and offline | Bengali | G:24 × 2, W:75 × 2 | Imitation of the performance and appearance of [74] |
| Synthetic OnOff SigHindi-75 [74] | On- and offline | Devanagari | G:24 × 2, W:75 × 2 | Imitation of the performance and appearance of [74] |
| Common nonpublicly available databases | | | | |
| 4NSigComp2010 [29] (Subset of GPDS960 [256] - Scenario 2) | Offline | Western | G:9600, F:1200, W:400 | 300dpi, BMP |
| GPDS-960 Signature DB [29, 78, 256] | Offline | Western | G:21144, F:26430, W:881 | Current database contains 881 users |
| GPDS-300 Signature DB [29, 78, 256] | Offline | Western | G:7200, F:9000, W:300 | This is a subset of GPDS960 database |
| Brazilian (PUC-PR) [89] | Offline | Western | G:6720, F:2280, W: 168 | |
| MCYT-330 [81, 191] | Online | Western | G:8250, F:8250, W:330 | Two sections. Fingerprints and signatures from the same donors |
| BiosecurID [79] | Online | Western | G:6400, F:4800, W:400 | Taken in four sessions |

*Number of G: genuine, F: forgery, D: disguised, W: writers.*

However, other scripts have been also considered in the literature. Although they are not as large as the West-ern corpuses, we can find databases in Malaysian [3], Arabic [6], Bengali [47, 74], Devanagari [74],Persian [98], Chinese [151], and Japanese [161]. It is worth pointing out that these databases are typically collected in the offline mode. It is expected that the development of real dynamic signa- ture databases in non-Western scripts will introduce stimulating and novel findings in automatic signature verification.

## 2.2. Considerations on Handwritten Signature Databases

Signature databases are supposed to gather realistic intraclass and interclass variability with all their specimens. Although many designers have made great efforts in approaching such reality in their corpuses, signatures are generally executed according to the willingness of the donors. Parental authorization for underage children, signing a large quantity of documents, signing an employment contract, signing in an awkward or uncomfortable way, or using different stylus devices to sign are a few situations where the context and the motivation to sign could dramatically affect the intraclass variability. In addition to failing to reproduce this intraclass realism in public signature databases, further considerations are highlighted in the following.

*2.2.1. General Considerations in Database Development.* Signature databases are not subject to quality assessment (QA) regarding the limitations in capturing intraclass and interclass variability. Further factors such as the emotions of the signers, the temporal evolution of children or adults [95,146], behavioral disorders, neurodegenerative diseases, or other cognitive impairments in fine mo-tor control [264] are not easily introduced in the current corpuses. Moreover, size limitation results in difficulty carrying out statistically meaningful validation and performance evaluation. Also, as in any biometric trait, data protection law is the main limiting factor to sharing the signatures freely among the research community, even though research is the only motivation. In addition, human mistakes, such as the mislabeling of specimens, could introduce a bias in the evaluation of algorithm performance.

*2.2.2. General Considerations in Offline Database Development.* Offline signatures are recorded from paper sheets. Often, static signatures can overlap the space available for another signature, e.g., signatures signed in a sheet, divided into several boxes [256]. In such a case, some signatures could cross the box borders or even overlap other signatures. Because of this effect, some signatures are manually discarded during the development of the database, although they should possibly be kept for the sake of realism. Low-resolution scanning can lead to poor image quality. Obviously, the higher the resolution, the more information can be extracted for image processing. Six hundred dots per inch (dpi) is the average resolution found in offline signature databases. There are further external factors that can limit the performance, such as noise in the imaging system, type of sensor used, blurring in the grayscale images, deficient ink deposition on the paper, and so forth.

*2.2.3. General Considerations in Online Database Development.* Some inconsistencies can be observed in the sampling rate. Although the data from a digitizing tablet may indicate that the sampling rate is, for instance, 100Hz, the timestamp, which is commonly used, may not reflect such a sampling rate during the whole signature acquisition. The velocity profile sometimes shows discontinuities during the transition from pen-down to pen-up. Such discontinuities could be at- tributed to inconsistencies in the device used, e.g., when the pen loses contact with the tablet.

*2.2.4. General Considerations in Online and Offline Database Development, Simultaneously Acquired.* To capture simultaneously online and offline signatures, a piece of paper is usually placed over the digitizing tablet (e.g., [74, 75]). Thus, the same signature is collected for the two modalities without scale or rotation variation between them. However, small movements between the paper and the tablet could introduce small and nonlinear distortions in some parts of the signa-tures. Because of such limitations in both offline and online signature databases, the accuracy of a system could be correlated to the database used. One possibility for avoiding incorrect conclusionsis to use as many signature databases as possible without adapting the ASV to each database. Thisshould lead to evaluating the accuracy of a system for different cases of intraclass and interclass variability, thus avoiding overfitting between a system and a database.

*Table 2.  Most Popular Algorithms for ASV Used during the Last 10 Years*

| Preprocessing | Techniques |
|---|---|
| - Enhancement | Background removal [48],** filtering noise [52, 53, 83],* printed text removal [78],** morphological operations [272],** ... |
| - Size normalization | Cropping [48, 98, 272],** fixed area [250],* [50, 209],** length equalization [206, 267],* alignments [250, 267],* [272],** ... |
| - Structural | Skeleton based [278],** ... |
| - No preprocessing | [82],* ... |

| Features | Techniques |
|---|---|
| - Function-features | Position [102, 246],* velocity [52, 53, 246],* acceleration [83],* pressure [56, 91],* pen direction  [227, 246],* length [235],* pen inclination [117, 149],* ... |
| - Global parameters | Number of strokes/lognormals [83, 248],* mathematical transformations (Wavelet [98, 200, 202, 203],* Radon [268],* fractal [84],** or others like total number, mean, and maximum of intra-/interstroke intersections, the number of x-axes, zero-crossings and the signature length [81, 91, 93, 154, 202, 246, 250]*), symbolic representation [7],* code vectors [234],* Sigma-lognormal based [83, 102],* ... |
| - Component oriented | Geometric [48, 229],** energy [182],** stroke thickness [129],** directions [223],** curvature [22],** graphometric [22],** ... |
| - Pixel oriented | Keypoints (SIFT [45],** SURF [163],** BRISK [61],** KAZE [184],** FREAK [159]**), run length [30],** texture [255],** quad-tree [232],** grid based [278],** shadow code [66, 223],** ... |

| Verifiers | Techniques |
|---|---|
| - Template Matching | Dynamic Time Warping (DTW) [52, 82, 83, 102, 140, 176, 234, 235, 236],* Manhattan [227],* Direct Matching Points [206],* Euclidean [98, 229],** ... |
| - Statistical measures | Mahalanobis [246],* [138],** membership functions [42],** [12],* [7],** cosine [208],**... |
| - Statistical models | Neural Networks (NNs) [144] and Deep Learning (Recurrent Neural Networks (RNNs) [4, 146, 247],*, Convolutional Neural Networks (CNNs) [268],* [11, 46, 109, 110, 139],** Deep Neural Networks (DNN) [220],** Deep Multitask Metric Learning (DMML) [240],** DCGANs [275]**), Hidden Markov Models (HMMs) [15, 72, 157, 251],* [48, 74],** Support Vector Machine (SVM) [103],* [54, 56, 76, 104, 198, 274],** Random Forest [203],* ... |
| - Structural | Decision Tree [209, 223],** graph models [205],** [263],*... |
| - Fusions | Ensemble of classifiers [21, 22],** [124],* template level [91],** score level [236, 250],* [91]*/**,... |

*\* Online signatures.\*\* Offline signatures.*

# 3. AUTOMATIC SIGNATURE VERIFICATION: A SUMMARY OF BEST PRACTICES

Automatic signature verification can be considered a two-class problem where the target is to decide whether a signature belongs to an enrolled individual or not. Although this challenge has been taken up in the past [123, 214, 218], many new algorithms and verification techniques have been reported in the last 10 years. ASV consists mainly of the preprocessing, feature extraction, and verification stages. For examples of the latest developments, see Table 2, which shows a non-exhaustive list of some of the most relevant algorithms for each stage.

At the preprocessing stage, the aim is to enhance the signature through image and signal processing techniques for static and dynamic signatures. At the feature extraction stage, the techniques can be divided into function-based features, which use the timing information in the signature; global parameters, which describe particular aspects of a signature (e.g., loops or baselines [142]); component-oriented features, which focus on particular characteristics, such as a pen-up or pen-down; and pixel-based techniques. At the verification stage, various strategies, some quite classical, are described in the recent literature, for example, dynamic time warping (DTW) [52, 82, 83, 102, 140, 176, 234–236] or support vector machines (SVMs) [54, 56, 76, 104, 198, 274]. However, deep learning [4, 11, 46, 109, 110, 139, 145, 220, 247, 275] seems to be one of the hot topics in ASV. It is worth pointing out that most biometric systems are nowadays based on deep learning approaches [23]. Furthermore, fusion approaches have also been popular in recent years, such as ensembles of classifiers [21, 22].

In the early research in signature verification, most of the systems reported in the literature were developed and evaluated for Western languages. Successively, Chinese and Japanese signatures were considered, and in recent years, along with the increasing size of the community working on this topic, a number of systems specialized in performing verification of signatures written in different scripts, such as Arabic, Persian, Hindi, and Bengali, have been described. Signature verification systems developed for different scripts can use similar approaches, but more frequently, they differ since they try to capture some specific characteristics of the particular script. In fact, although the general model underlying the signature generation process is invariant in terms of cultural habits and language differences between signers, the enormous diversity in the signatures of people from different countries has led to the development of script-specific solutions. For instance, in some countries, the habit is to sign with a readable, written name, whereas in other countries signatures are not always legible. As the need for cross-cultural applications increases, it is more and more important to evaluate both the extent to which the personal background affects signature characteristics and the accuracy of the verification process. In order to emphasize the efforts of the scientific community in adapting systems to different scripts, this section is organized on a script basis.

First of all, Table 3 gives an overview per script of the best system performance achieved with random (RF) and skilled forgeries (SK), taking into account the algorithm used, the number of genuine reference signatures (#T) per user, and the databases most used in the previous decade.

### 3.1. Western Signature Verification Systems

For *offline signatures,* we can find either writer-dependent proposals [165], which build a mathematical model per writer, or writer-independent proposals [112], which develop a unique modelin the ASV. One was a set of global features based on the boundary of a signature, such as its totalenergy, the vertical and horizontal projections, and the overall box size in which the signature is contained [181]. Also, in [229], a system based on simple geometric features is presented, with results similar to some others [123, 214, 218]. In this work, a centroid feature vector was obtained, foreach user, from a set of genuine samples. The centroid signature was then used as a template for verifying a claimed identity, through a Euclidean-distance-based approach and by using a localizedthreshold.

Other authors focused on the global image level and measured the gray-level variations in thesignature strokes by identifying statistical texture features [255]. In [223], a multifeatured extration method was proposed that allows one to extract a range of global to very local features byvarying the scale of a virtual grid used to partition each signature image. The signature can alsobe described by a quad-tree structure and an artificial immune recognition (AIR) system for verification [232] or through interest points such as SIFT [45], SURF [163], BRISK [61], KAZE [184], or FREAK [159]. Additionally, fuzzy membership functions are still used in signature verification. Forexample, in [7], a fuzzy similarity measure is used to classify a symbolic representation of offline signatures, and this method also achieves a competitive performance.

In recent years, there has been a growing interest in feature representation learning: insteadof relying on handcrafted feature extractors, feature representations are learned directly from the signature images by applying deep learning techniques, as seen in Table 2, such as convolutional neural networks (CNNs), deep multitask metric learning (DMML) [240], or deep convolutional generative adversarial networks (DCGANs) [275]. For example, a Siamese network based on twin CNNs is developed in [46]. CNN [11] is applied to the learning of features in a writer-independentformat in [109]. Here the model is used to obtain a feature representation for another set of users, which feeds into writer-dependent classifiers. Such work was extended in [110] by a novel formulation of the problem that includes applying a knowledge of skilled forgeries during the featurelearning process. A similar approach based on deep metric learning [240] can be found in [220].Comparing triplets extracted from two genuine and one forged signature, the proposed systemlearns to embed signatures into a high-dimensional space, in which the Euclidean distance is used as a similarity metric. In [240], a classification method based on DMML is studied. This approachmakes use of knowledge from the similarities and dissimilarities between genuine and forged samples in order to learn a distance metric that is applied to measure the similarity between pairs of signatures.

*Table 3. Performances (%) for Some Published ASV Systems in Different Scripts*

| WESTERN | | | | | | | | | |
|---|---|---|---|---|---|---|---|---|---|
| **Database** | **Method** | **#T** | **RF** | **SK** | **Database** | **Method** | **#T** | **RF** | **SK** |
| GPDS-300 *Offline* | Curvelet+OC-SVM [104] | 4 | - | AER:8.70 | MCYT-100 *Online* | VQ+DTW [234] | 5 | - | EER:1.55 |
| | Learning Feat.+CNN [110] | 5 | - | EER:2.42 | | WP+BL DTW fusion [236] | 5 | - | EER:2.76 |
| | HOG,DRT+DMML [240] | 5 | EER:2.15 | EER:20.94 | | Histogram+Manhattan [227] | 5 | EER:1.15 | EER:4.02 |
| | LBP, LDP+SVM [48] | 2 | EER:1.43 | EER:21.63 | | Sigma-Lognormal+DTW [83] | 5 | EER:1.01 | EER:3.56 |
| | Quad-tree+(AIR) [232] | 10 | - | EER:9.30 | | Symbolic representation [108] | 5 | EER:1.85 | EER:5.84 |
| | Grid Feat.+SVM [278] | 12 | - | EER:3.24 | | GMM Feat.+DTW [235] | 5 | - | EER:3.05 |
| CEDAR *Offline* | Learning Feat.+CNN [110] | 4 | - | EER:5.87 | SVC2004 *Online* | Function-based+RNN [145] | 10 | - | EER:2.37 |
| | Surroundedness+NN [144] | 1 | - | AER:8.33 | | VQ+DTW [234] | 5 | - | EER:2.73 |
| | Quad-tree+(AIR) [232] | 10 | - | EER:18.15 | | DCT+Sparse Rep. [149] | 10 | EER:0.45 | EER:5.61 |
| | Grid Feat.+SVM [278] | 10 | - | EER:3.02 | | Fuzzy modeling [12] | 5 | EER:5.49 | EER:7.57 |
| MCYT-75 *Offline* | Learning feat.+CNN [110] | 5 | - | EER:3.58 | SUSIG-Visual *Online* | DTW+Lineal Classifier [140] | 5 | EER:4.08 | EER:2.10 |
| | HOG,DRT+DMML [240] | 5 | EER:1.73 | EER:13.44 | | Sigma-Lognormal+DTW [83] | 5 | EER:1.48 | EER:2.23 |
| | LBP, LDP+SVM [48] | 2 | EER:0.69 | EER:16.03 | | Target-wise+DTW [52] | 1 | EER:1.55 | EER:6.67 |
| | Grid Feat.+SVM [278] | 10 | - | EER:4.01 | | Histogram+Manhattan [227] | 5 | EER:2.91 | EER:4.37 |
| CHINESE | | | | | | | | | |
| **Database** | **Method** | **#T** | **RF** | **SK** | **Database** | **Method** | **#T** | **RF** | **SK** |
| Own Database *Offline* | Segm.+Similarities [131]† | 3 | FRR:13.23 FAR:0.03 | FRR:13.23 FAR:29.75 | Sigcomp2011 *Online* | FHE-based Feat.+RF [203] | 80%* | - | EER:7.45 |
| | | | | | | FHE-based Feat.+RF [202] | 80%* | - | EER:5.98 |
| Own Database *Offline* | Segm.+Weig. Sim. [129]‡ | 5 | FRR:3.32 FAR:0.09 | FRR:3.32 FAR:21.20 | | Legendre poly.+RF [201] | 80%* | - | EER:8.93 |
| JAPANESE | | | | | | | | | |
| **Database** | **Method** | **#T** | **RF** | **SK** | **Database** | **Method** | **#T** | **RF** | **SK** |
| Own Database *Off/online* | Off-/On-line ASV [128] | 3 | - | ACC:89.65 | Own Database *Offline* | Gradient+Mahalanobis [136] | 3 | - | ACC:92.25 |
| Own Database *Off/Online* | Off-/On-line ASV [137] | 3 | - | ACC:95.47 | SigWiComp2013 *Off/Online* | Off-/On-line ASV [169] | 11 | - | EER:5.51 |
| ARABIC | | | | | | | | | |
| **Database** | **Method** | **#T** | **RF** | **SK** | **Database** | **Method** | **#T** | **RF** | **SK** |
| Own database *Offline* | Geometrical+DTW [70] | N/A | FRR:8.00 FAR:13.00 | - | QU-PRIP *Online* | Function-based+DTW [63] | 4 | EER:0.3 | - |
| Own database *Offline* | Mult. feat+fuzzy [42] | N/A | ACC:98% | - | | | | | |
| PERSIAN | | | | | | | | | |
| **Database** | **Method** | **#T** | **RF** | **SK** | **Database** | **Method** | **#T** | **RF** | **SK** |
| UTSig *Offline* | Geom. Features+SVM [241] | 12 | - | EER:29.71 | NDSD *Online* | Function-based+SVM [267] | 3 | - | EER:5.15 |
| | HOG,DRT+DMML [240] | 12 | - | EER:17.45 | | RT+CNN [268] | 5 | - | EER:2.07 |
| INDIC | | | | | | | | | |
| **Database** | **Method** | **#T** | **RF** | **SK** | **Database** | **Method** | **#T** | **RF** | **SK** |
| BHSig260 (Bengali) *Offline* | Contour Feat.+SVM [193] | 12 | - | AER:12.33 | OnOffSigBen-gali75 *Online* | Function-based+DTW [74] | 2 | EER:0.27 | - |
| | Texture Feat+SVM [51] | 2 | EER:1.78 | EER:10.67 | | Histogram+Man [74] | 2 | EER:8.19 | - |
| | Texture Feat+NN [192] | 8 | - | EER:33.82 | | | | | |
| BHSig260 (Hindi) *Offline* | Texture Feat+NN [192] | 8 | - | EER:24.47 | OnOffSigHin-di75 *Online* | Function-based+DTW [74] | 2 | EER:0.41 | - |
| | Texture Feat+SVM [51] | 2 | EER:1.34 | EER:11.88 | | Histogram+Man [74] | 2 | EER:9.77 | - |

EER: equal error rate, AER: average error rate, ACC: verification accuracy. *Percentage of available genuine signatures per user.

FAR was also reported for simple forgeries: †FAR: 9.28, ‡FAR: 16.52.

Ensemble strategies to classify signatures have also been investigated where multiple classifiersare trained on the same dataset. For instance, an ensemble of classifiers is proposed in [22]. These are based on graphometric features to improve classification performance in the offline case. The ensemble was built using a standard genetic algorithm that selects a subset of models based on different fitness functions. An analogous approach in [21] used multiple hidden Markov models for learning signatures at different levels of perception.

One of the main constraints for commercializing an ASV is the required number of genuine sig-natures to be taken per user. This leads to the necessity to adapt and evolve these systems [123]. Toaddress this problem, the use of one-class SVM (OC-SVM) is proposed in [104, 112]. OC classification attempts to classify just OC of objects (only the genuine signature, in this particular case), thusdistinguishing them from all other possible objects. In [66], a hybrid writer-independent/writer-dependent system is proposed: when a user is enrolled in the system, i.e., when only few signa-tures are available, a writer-independent classifier is used to verify the signature. However, once the number of collected genuine samples passes a threshold, a writer-dependent classifier replacesthe previous one for the specific user. Unfortunately, the low number of genuine samples takenrepresents a critical drawback of this strategy. Instead, in [48, 51, 73], augmenting the training signatures by generating synthetic offline handwritten signatures is suggested for neuromotor-inspired models.

For *online signatures*, in recent years, an increasing effort has been devoted to applying functionfeatures based on pressure and force. In [261], for example, specific devices were developed to capture these functions directly during the signing process. The directions of the pen movements and the pen inclinations have created a growing interest [117, 200], as have the turning angle rep- resentations [19]. In [122], the authors proposed a novel approach for online signature verificationbased on the directional analysis of velocity-based partitions of the signature. In [149], the ASV was based on the discrete cosine transformation (DCT) and a sparse representation by fixing a certain number of coefficients. A similar approach was studied in [200] on the basis of a discrete 1D wavelet transform.

Concerning the verification phase, whatever the technique adopted, the need to improve verification performance has led researchers to investigate multiexpert approaches [124] for combining sets of verifiers based, for example, on global [81, 167] and local strategies [178]. Up to now, several multiexpert systems have been proposed in the literature, based on abstract-level, ranked-level,and measurement-level combination methods as well as on serial, parallel, and hybrid topologies. Moreover, fusion schemes at the score level or feature level are also popular nowadays [91, 236, 250].

Proposals for online ASVs, designed on the basis of a DTW [52, 82, 83, 102, 140,

176, 234–236]have often been used over the years [123, 214, 218]. It seems that such elastic matching copes adequately with the problem of specimens with different lengths and internal variations. Some recent approaches focus on the normalization of the DTW (e.g., [52, 82, 140, 176]), while others focus on fusing stable local multidomains in a signature [206] or by the exploitation of the optimization matrix [236], which can be calculated by using Euclidean or Cityblock distances, among others. Moreover, DTW is commonly used in combination with vector quantization [234], GMM features [235], Sigma-Lognormal-based features [83, 102], or a kernel function such as SVM [103]or OC-SVM [173].

A scheme based on fuzzy measures of signatures segments was designed in [12]. Also in [39, 40], a neuro-fuzzy classification stage of partitioning features (vertical [39] and horizontal [40]) was studied. A representation of online signatures by interval-valued, symbolic parameter-based features is studied in [108]. These authors achieved the best verification results when a writer- dependent threshold was adopted for distance-based classification. A similar scheme based on interval-valued symbolic features can be found in [107], which was based on both a symbolic and conventional representation of online signatures.

The measure, based on local density estimation by a hidden Markov model (HMM), can be usedto ascertain whether a signature contains sufficient information to be successfully processed by a signature verification system. Statistical models like HMM [157] or histogram features [227] are nowadays also used in signature verification. However, the popularity of schemes based on RNN is emerging [4, 145, 247].

Finally, notwithstanding the numerous measures that have been proposed so far, performance evaluation still remains a critical and complex task. For example, as shown in [10], a significantdegradation in the verification of online signatures can be observed as the quality of the forgeries used increases.

## 3.2. Chinese Signature Verification

Much work has been carried out in the field of Chinese signature verification, for both offline and online systems.

For *offline signatures,* a proprietary dataset was collected in [129–131]. In [130], it is proposed that, after preprocessing, the signatures are divided into chained segments, represented by seven geometric features. A similarity measure is then used for computing the FAR and FRR for genuine,random, and simple forgeries. Signature segmentation in different regions of the script was car- ried out in [129, 131]. Then seven features, grouped into three types (geometric centers, segment tracks, and stroke thickness in each segment), were extracted. The experiments were performed with and without applying different thresholds and weighting factors. The system generally had problems with skilled forgeries as compared to simple and random forgeries and genuine signa- tures. A system for identifying offline Chinese signatures was presented in [197]. It was based on foreground and background features. The authors used a database of 1,120 (640 English and 480 Chinese) signatures and report an identification rate of about 97%. In [147], the authors summarizethe major issues, challenges, and future directions regarding forensic document and handwriting examination in China.

For *online signatures,* orthogonal series expansions of Legendre polynomials and wavelet de- composition were used as features in [203], followed by a random forest (RF) classification. Improvements were found in [202] through an RF classifier and global features (e.g., total time, pen-up/down duration, maximum pressure time, etc.) and wavelet representation of the time functions. Experiments in both of these studies were conducted with the Sigcomp 2011 Chinese dataset.An interesting aspect of each study, i.e., [202, 203], was the use of relevant features as identified by forensic handwriting examiners (FHEs). A different proposal applied to the same dataset is found in [201]. In this case, the authors used Legendre polynomials from function-based features and an RF classifier. A methodology in [148] was focused on verifying online signatures collected on mobile devices. The features used included xy-coordinates, pressure, and contact area. User- dependent classification was performed using four classifiers, i.e., SVM [274], logistic regression,AdaBoost, and RF. Adaboost performed the best on a small dataset of 42 persons collected forthis study. In [37], function-based features and template matching were applied to a proprietary Chinese signature dataset of 90 subjects.

In [183], online Chinese signatures from Sigcomp 2011 were converted to offline [56]. The authors then carried out a combined signature verification using grayscale gradient features and Mahalanobis distances for offline signatures and dynamic programming matching with function- based features for online signatures.

### 3.3.Japanese Signature Verification

It is common practice in Japan that people who possess their name seals/stamps prefer using them for the purpose of authenticating documents instead of using handwritten signatures.[1] Becauseof such a tradition, handwritten signatures are less practiced there. This leads to lack of stability and individuality in handwritten Japanese signatures. For this reason, only a limited number of approaches have been recently reported for Japanese signature verification.

Apart from specific benchmarks [161], we found in the literature combined approaches for *offline* and *online* Japanese signature verification. An in-house, online dataset of 45 individuals was collected for the work presented in [138]. The online data were converted to offline by using the velocity profile. The verification system consisted of gradient features and a Mahalanobis distance.The authors obtained a FAR and FRR of 91.87% and could improve the accuracy of the system by4.9% when signatures from other individuals were used. In [128], the authors used two

ASVs for online and offline data, which were generated in a way similar to [138]. The static ASV was basedon gradient features and a Mahalanobis distance, whereas the dynamic ASV was function based and used dynamic program matching. Finally, both techniques were combined with an SVM, thus obtaining promising results on a dataset of 44 individuals. A similar strategy was presented in [137]with a dataset of 19 writers. The experiments included cases where the signatures were first segmented and cases where the complete signatures were used during verification. Also, in [136], a static ASV [138] is presented by using a nonpublic dataset of 19 individuals. Again, data were collected online and converted into offline [138]. Finally, in [169], an off- and online combined systemis presented by using as input the online Japanese signatures from SigWiComp 2013 [161].

### 3.4. Arabic Signature Verification

Arabic communication is full of gesture. This is also seen in handwriting, especially in signatures.Consequently, an ASV has to pay attention to this peculiarity.
For *offline signatures,* in [127], the authors present a two-stage system for offline Arabic signature identification and verification. The first stage applies image morphology and focuses on the extraction of various global and local geometric features. Verification is performed at the second stage. Evaluation is conducted on a dataset of 330 handwritten Arabic signatures (220 genuineand 110 forged). The system achieved an identification rate of 95% and a verification rate of 98% when trained on genuine signatures (132 samples) only and was tested on both genuine and forgedsignatures (198 samples). In [6], two datasets were introduced for offline and online Arabic sig- nature verification. The offline dataset contains 12,000 signatures from about 1,000 writers, and the online/offline data set contains 1,750 signatures from 190 writers. The data were collected from individuals belonging to different age groups, nationalities, genders, and academic qualifications. Geometrical features, such as diagonal and polar measures, were extracted from image-based signatures and a DTW was the ASV used in [70]. Authors built their own database with genuine signatures from 10 volunteers. Additionally, a proposal based on multiple feature fusion with fuzzymodeling can be found in [42]. Similarly, in [164], a system was proposed based on various geo- metrical and grid features. This has been evaluated on a small dataset of 360 signatures (60 writerswith six signatures each) and gave a false rejection of less than 1%.
For *online signatures,* function-based features and a DTW was the strategy used in [63]. Authorsused 43 signatures from the Arabic Off/On QU-PRIP dataset [6].

### 3.5. Persian Signature Verification
Persian and Arab scripts are very similar, though in Persian, people generally draw shapes, andthis brings new challenges to signature verification [98]. Recent approaches to Persian signature verification are reviewed below.

For *offline signatures,* several approaches have been published utilizing private databases. For example, [98] presented a system based on a discrete wavelet transform (DWT) and the Euclideandistance for matching. A similar approach was published in [99], where the authors created a reference pattern of users by fusing registered signatures. In [139], the authors used deep learning,i.e., CNN, to conduct the random forgery experiment. Additionally, a method based on estimating fractal dimensions of Persian offline signatures can be seen in [84]. More recently, authors proposeto use geometrical features based on fixed-point arithmetic followed by an SVM as a classifierin [241]. Experiments were conducted with the UTSig database in terms of EER using, as nontargetclasses, random and skilled forgeries. The same authors obtained better results in [240] with the same dataset by using a DMML method.

For *online signatures,* several techniques based on DTW with multivariate autoregressive(MVAR) modeling and a multilayer neural network were presented in [277]. The authors vali- dated their system with their own dataset, which was later known as the Zoghi dataset. A system based on global information from online signatures, adjusted with an adaptive network-based, fuzzy inferencing algorithm, was introduced in [44]. This work introduced another dataset, namedthe Dehghani dataset, and considered both random and skilled forgeries. In [267], a system based on robust features and an SVM was developed and evaluated on three datasets of Persian signa- tures [44, 277] and on the NDSD. Most recently, a Radon transform (RT) and a CNN was usedin [268]. The authors evaluated the system on three datasets (Dehghani [44], Zoghi [277], and NDSD [267]) of Persian signatures.

## 3.6. Signature Verification in Indic Languages

India is a country rich in languages. In spite of the fact that there are 122 main languages in India and 1,599 secondary languages [259], Hindi and Bengali are the official languages. Hindi is expressed in Devanagari script. This explains why the majority of papers in this area consider onlyHindi and Bengali scripts. In this subsection, we summarize the contributions reported in the last few years.

For *offline signatures,* an ASV that uses Zernike moments and gradient features and SVM for classification for Hindi signatures was introduced in [200]. Competitive error rates in terms of FAR and FRR were obtained with a subset of the BHSig260 [192] corpus. Similarly, [193] presenteda system utilizing contour features, such as the intersection and endpoint chain code methods, along with an SVM classification for Bengali signature verification. The system was evaluated on a Bengali dataset [192]. Another ASV based on texture features and SVM was applied to both Hindiand Bengali signatures in [51, 74]. Thus, local binary patterns (LBPs) and uniform LPB strategies were used for parameterized Bengali and Hindi signatures in [192]. The classification was carried out through an NN technique for Hindi and Bengali signatures separately. By using the same dataset, promising results were found in [7] using a symbolic interval representation of signatures and fuzzy similarities. In this case, the authors performed experiments without separating Ben-gali and Hindi signatures. In [46], deep learning was considered for both languages and obtained competitive performance for Bengali signatures.

For *online signatures,* little work has been published. In [74], an ASV based on function features and DTW was used for OnOffSigBengali-75 and OnOffSigHindi-75 datasets. Only random forgery experiments were performed. Also, an in-house implementation of the algorithm proposed in [227] was tested on the same datasets as used in [74].

*Table 4. Summary of Signature Verification Competitions Organized in the Last 10 Years*

| Competition | Datasets | Verification Tasks | Metrics | Best Results |
|---|---|---|---|---|
| **BSEC2009** | DS2-382(digitizer tablet):382(A) DS3-382(PDA):382(A) | T1 (DS2-382): Online T2 (DS3-382): Online | (EER) | T1: 0.51(RF), 2.20(SF) T2: 0.55(RF), 4.97(SF) |
| **SigComp2009** | 1920(S), 100(A), 12(G)/(A), 6(F)/(A) | T1: Online T2: Offline T3: Combined | (EER) | T1: 2.85(SK) T2: 9.15(SK) T3: 8.17(SK) |
| **4NSigComp2010** | Scenario 1: 2(A), 354(S), 115(G), 194(F), 27(D) Scenario 2: GPDS960 data (Author 301 to 960) | T1 (Scenario 1): Offline T2 (Scenario 2): Offline detection of (RF) and (SK) | (EER) (OE) | T1: 55.0 T2: 8.94 |
| **ESRA2011** | DS2-382 (digitizer tablet): 382(A) DS3-382 (PDA): 382(A) | T1: Impact of forgery quality T2: SK categorization | (EER) | T1 (DS2): 2.73(BQ), 2.85(GQ) T1 (DS3): 6.05(BQ), 7.15(GQ) T2 (DS2): 3.32(BQ), 4.31(GQ) T2 (DS3): 1.67(BQ), 2.43(GQ) |
| **SigComp2011** | Dutch: 4653(S), 128(A), 3162(G), 1491(F) Chinese: 2544(S), 40(A), 946(G), 1598(F) | T1: Chinese offline T2: Dutch offline T3: Chinese online T4: Dutch online | $(\widehat{C}_{llr}^{min})$ | T1: 0.69 T2: 0.08 T3: 0.22 T4: 0.12 |
| **4NSigComp2012** | 450(S), 3(A), 113(G), 273(F), 64(D) | T1: Offline verification | $(\widehat{C}_{llr}^{min})$ | T1: 0.43 |
| **SigWiComp2013** | Dutch: 1242(S), 27(A), 270(G), 972(F) Japanese: 4836(S), 62(A), 2604(G), 2232(F). | T1: Dutch offline T2: Japanese offline T3: Japanese online | $(\widehat{C}_{llr}^{min})$ | T1: 0.64 T2: 0.33 T3: 0.74 |
| **SigWIcomp2015** | Bengali: 540(S), 10(A), 240(G), 300(F) Italian: 728(S), 50(A), 479(G), 249(F) German: 750(S), 30(A), 450(G), 300(F) | T1: Italian offline T2: Bengali offline T3: German online | $(\widehat{C}_{llr}^{min})$ | T1: 0.02 T2: 0.03 T3: 0.29 |

Signature (S), Genuine Signatures (G), Disguised Signatures (D), Forgeries (F), Random Forgeries (RF), Simple Forgeries (SF), Skilled Forgeries (SK), Number of Authors (A), Bad-Quality Forgery (BQ), Good-Quality Forgery (GQ), Overall Error (OE), Equal Error Rate (EER), Minimum Cost of Log Likelihood Ratios $(\widehat{C}_{llr}^{min})$.

## 4.   4 COMPETITIONS: THE STATE OF THE ART

In Section 3, we presented the best results obtained when authors evaluated their own systems under particular experimental protocols. These results were not directly comparable. Indeed, it could be said that every contribution has a particular experimental protocol that makes them completely incomparable in terms of accuracy. Nevertheless, in the pattern recognition (PR) community, several competitions have been organized during the last 10 years as part of the most relevant conferences in the field such as ICDAR and ICFHR. These competitions established common benchmarks and common challenges to test third-party ASVs.

 These external comparisons give a more objective evaluation of the state of the art in automatic signature verification and define the goal for publishing results on new systems. A comprehensive summary of the best ASV systems from these competitions is provided in Table 4.

 At the ICDAR 2009 Signature Verification Competition (SigComp 2009) [28], a new challenge attracted the participation of 24 systems. It consisted of designing one reference signature system for skilled forgery detection. The systems were tested to cope with offline and online signatures Additionally, only one system combined both online and offline signatures in a single system. The evaluation was based on equal error rates (EERs).

 The BioSecure Signature Evaluation Campaign (BSEC 2009) [119] was the next competition and was organized during the ICB 2009. The organizers developed an online signature database acquired from a Wacom digitizing tablet and another database acquired on a PDA HP [190]. Both included the same number of 382 signers. Twelve systems participated in this competition and were evaluated in two tasks on skilled and random forgeries on the two datasets. The BioSecure Evaluation of Signature Resistance to Attacks (ESRA 2011) was organized in 2011 using data from BSEC 2009 with 13 participating systems. This competition focused on assessing online ASV systems in terms of the impact of forgery quality (good or bad) on the systems' performance (Task 1) and on testing the systems using various categories of skilled forgeries (Task 2) [115]. The results for the winning systems are presented in Table 4.

Note that FHEs point out various issues related to these signature verification competitions, for example, verifying one questioned signature against one reference signature (such as in SigComp 2009) as well as focusing exclusively on online data. This makes sense in terms of developing future online commercial applications of ASV, but the duty of the FHE [152] is the verification of only offline signatures. Furthermore, these competitions did not look into different genres/categories of signatures as encountered by forensic examiners, and the results were not reported in a way required in forensic casework [62]. In summary, the underlying motivation of organizing these competitions differed from the FHEs' requirements. For this purpose, FHEs joined with the pattern recognition community to collectively organize a series of signature verification competitions targeting the basic signature verification requirements of FHEs.

Two Signature Verification Competitions (4NSigComp 2010) were proposed during the ICFHR 2010: Scenario 1 [152] and Scenario 2 [29]. Scenario 1 [152] was proposed to detect genuine signatures, skilled forgeries, and disguised signatures, which are personal signatures falsely executed by the original writers. Seven verification systems participated, and their results were compared with FHE evaluations using the same set of signatures. It was observed that the systems were not able to satisfactorily detect disguised and skilled forgeries at the same time. In fact, when disguised signatures were included in the evaluation set, the performance worsened significantly [152]. An interesting fact, however, was that the performance of the competing ASVs was really close to the performance obtained by the FHEs [162]. In Scenario 2 [29], 10 offline ASVs were evaluated for detecting random and skilled forgeries at the same time. The results were given in terms of overall error (OE). The best-performing system achieved an OE of 8.94%.

In the Signature Verification Competition (SigComp 2011) [151] held in the ICDAR 2011, a close approximation to the methods of the forensic community was attempted through the performance evaluation by EER and the use of the minimum cost of the log-likelihood ratios (LRs),$c_{llr}^{min}$. A unique feature of SigComp 2011 was that the 13 participating systems were required to report continuous scores, which could be converted into LRs or Log LRs (LLRs) by various calibration procedures. The cost of log-likelihood ratios $C_{llr}$ and the minimum possible value of $C_{llr}$, i.e., $C_{llr}^{min}$, were then computed. The smaller the value of $C_{llr}^{min}$, the better the system. In addition to these measures, forensic-like databases were used: online and offline Western and Chinese corpuses. A single system able to deal with all cases was not found. The best results for different tasks are reported in Table 4. One interesting finding was that the best minimal cost of log-likelihood ratio did not imply the best EER. This is attributed to the sensitivity of $C_{llr}^{min}$ to a few important errors during the verification process.

In ICFHR 2012, $C_{llr}^{min}$ was used to rank the systems in the competition (4NsigComp 2012) [150]. In this competition, genuine, disguised, and forged signature detections were evaluated. Five systems participated and the winning system reached a $C_{llr}^{min}$ of 0.43. When disguised signatures were removed from the test set, once again the same system succeeded with a $C_{llr}$ of 0.36.

At ICDAR 2013, the competition (SigWiComp 2013) [161] focused on online and offline signature verification against skilled forgery detection. Three subcorpuses designed by forensic experts were used: a Western offline (Dutch signatures) and an online and offline Japanese dataset, the latter being the offline Japanese counterpart generated directly from the online data. Thirteen systems participated, some with optimized versions for different tasks. One interesting finding in this competition was that the $C_{llr}^{min}$ results tended to be more sensitive to a few large errors than the EER. It must be remembered that this is the clue to using such a metric in forensic casework, where omitting large errors would necessarily result in making a dramatically wrong decision in a court of law.

The SigWIcomp 2015 [160] signature verification competition was held during ICDAR 2015. The tasks dealt with offline signature verification using Western (Italian) and Indian (Bengali) signatures as well as Western online specimens (German). Only the $C_{llr}^{min}$ results were presented. Thirty systems participated for different verification tasks. One of the main outcomes from this competition was the evaluation of useful ASVs for forensic daily practice, and this was suggested as the future direction to follow. This is a questionable move, considering the widespread availability of handheld devices with their new handwriting digitizing ability and their increased computing power, which allows an easy implementation of a built-in ASV. We briefly come back to this point in the conclusions.

Many of these competitions were organized by the Netherlands Forensic Institute in conjunction with pattern recognition researchers. This has led to the use of signatures found in typical forensic casework and the development of systems useful for both communities. Furthermore, multiscript tasks have also gained attention in recent competitions. In fact, it can be seen that participants optimized their systems for each task in SigWIcomp 2015 [160]. A more practical and challenging scenario could consider different intrawriter scripts for the training and evaluation set in order to prove fairly the performance of the state-of-the-art technology for several scripts at the same time. Also, despite its necessity in commercial transactions and applications, the random forgery scenario has become of low interest in many of the latest competitions, which focus mainly on skilled forgeries. However, recent papers still show results for the random forgery scenario [43, 278].

## 5.  FORENSIC ASPECTS

FHEs perform signature comparison along with signature verification as they address the identity of a signer [152]. During the examination, an FHE compares a questioned signature against various known specimens. An FHE assumes that different writers do not write the same way and every piece of handwritten signature they write is essentially influenced by the so-called intrawriter and interwriter variations. In order to show that the questioned and specimen signatures were written by the same writer, an FHE must establish that the degree of variation is more consistent with the intrawriter variations than with the interwriter variations. However, in some cases, e.g., tracing, the intrawriter variations are so negligible that an FHE can consider it with high likelihood [182] as a simulation. After an examination, an FHE tends to express an opinion about the observations supported by one of the following hypotheses [17, 85, 86, 175]: (1) The questioned signature is genuine, i.e., naturally written by the original writer in his or her own style. (2) The questioned signature is not genuine (is a simulation), i.e., unnaturally written by some writer other than the specimen writer—the writer tried to make a false copy of the specimen signature. (3) The questioned signature is naturally written by a person in his or her own style but falsely claiming to be some other person. (4) The questioned signature is not genuine, i.e., unnaturally written by the writer of the questioned specimen—the writer tries to make his or her signatures look like a forgery attempt (referred to as "disguise" behavior). The purpose could be for the later denial of writing the signature, in which case it is also termed as "auto-simulation." Note that both genuine and disguised signatures are written by the author of the specimen but with different intentions. A signer produces a genuine signature to be identified when needed, whereas a disguised signature is written with the purpose of making it look like a forgery for a possible later denial [152, 175]. (5) The questioned signature is unnaturally written by the specimen writer under the influence of internal or external factors, e.g., illness, alcohol.

The results of signature comparisons carried out by FHEs are subjective. They usually construct weighted hypotheses. FHEs often express their opinion on a five- or nine-point scale ranging from "identification" of a definite match between the questioned and specimen signatures to "elimination." The latter represents the full confidence of the FHE that the questioned and specimen signatures are not written by the same writer.

However, some automatic tools have recently been developed specifically to support FHEs in their daily casework, e.g., FLASH ID [260] in 2013, iFOX [243] in 2013, and D-Scribe [230] in 2013. It is, however, interesting that with respect to signature authentication, FHEs have traditionally made very limited use of such automated tools. One reason for this is that many of these tools have been designed to perform comparison tasks and present results in a form that FHEs are not comfortable with. Furthermore, some tools have certain drawbacks that affect the possibility of their usage in the real world of forensic casework. For example, CEDAR-FOX, introduced by the Center of Excellence for Document Analysis and Recognition for supporting a semiautomated analysis for handwritten input, has an output based on the assumption that the appearance of each occurrence of a character is independent of the next. If this assumption were true, there would be no point in comparing handwriting with the purpose of source attribution. The behavior of the system differs from that which would be expected in many forensically relevant scenarios [257].

The Forensic Language-Independent Analysis System for Handwriting Identification (FLASH ID) [260] identifies writers by analyzing grapheme topology and geometric features. The system iFOX [243] (interactive FOrensic eXamination) and D-Scribe [230] have been introduced more recently to assist in handwriting interpretation through automation. Unfortunately, none of these systems is in such widespread use that its utility can be fully established.

## 6. RECENT PROGRESS IN AUTOMATIC SIGNATURE VERIFICATION

While Section 3 gave an overview of novel ways to solve a well-known challenge, i.e., automatic signature verification, various hot topics are discussed in this section. These bring new challenges to the research community as well as providing a source of opportunities and accomplishments. As such, we focus this section on ASV systems that process signatures in more than one script (Section 6.1), on the developed stability and complexity measurements on signatures (Section 6.2), on international standards and systems that operate independently of the data-acquiring device (Section 6.3), on the algorithms proposed to encrypt a person's signature (Section 6.4), on signature-based e-health applications to aid objective medical decision making (Section 6.5), on generative models to reproduce synthetically handwritten signatures (Section 6.6), and on touchscreen devices (Section 6.7).

## 6.1 Multiscript Signature Verification Systems

Multiscript refers to a scenario where signatures in several different scripts (i.e., using letters or characters in different languages) can be verified by using the same ASV. Despite not being extensively studied [196], multiscript is a common scenario for international commercial transactions, international companies, or international security for forensic paradigms. In all these cases, signatures are typically executed in the original script and can be aggregated to a common database.

To evaluate an ASV under multiscript conditions, proposals in the literature usually combine several single-script databases. As an example, in [199], the authors evaluated an ASV after merging databases in several scripts. These authors presented two offline signature verification experiments: first without initial script identification and second with initial script identification for English, Hindi, and Bengali. The system developed for this study utilized chain code features and signature gradients along with support-vector-based classification. Results indicated that initial script identification improves the final verification results by almost 4%.

Most recently, [43] presented a study on the evaluation of multiscript versus single-script signature verification scenarios and a technique for generating a multiscript offline signature database from several single-script databases. The study considers eight offline public signature datasets, available in five different scripts (Western, Bengali, Devanagari, Chinese, and Arabic). The authors showed that similar results are obtained when merging the datasets or keeping them singular, thereby implying multiscript verification as a generalized problem.

In [194], gradient features and chain code features were fused at the template level for verification of Bengali and English signatures. High accuracies of 99.41%, 98.45%, and 97.75% were obtained by studying different techniques. In [197], English and Chinese signatures were combined. In this multiscript scenario, the authors applied a foreground and background technique for the identification stage. They achieved an accuracy of 97.70% in identification.

Only Bengali, Devanagari, and Western scripts were evaluated in [195]. From their analysis, these authors observed large errors in the classification of Bengali and Devanagari signatures. Their follow-up work in [198] consisted of using a modified gradient feature and an SVM for identification and verification. Some 66% of the signatures from each script were used for training and 33% for testing. The system achieved a script identification accuracy of 98%, and the false acceptance and false rejection rates for Hindi signature verification were 8% and 4%, respectively; for English signatures, they were 12% and 10%, respectively. LBP and Uniform LBP features were studied in [192] for offline verification of signatures written in two Indic scripts, which were Bengali and Hindi. The presented system was evaluated on BHSig260 as well as on the publicly available GPDS-100 signature dataset. The system achieved an EER of approximately 34% on Bengali, 24% on Hindi, and 33% on GPDS-100 data.

Furthermore, three datasets from three languages (Dutch, Chinese, and Japanese) were combined to create a training dataset in [169]. The authors found competitive results when an ordered sampling method based on OC-SVM was used for genuine samples of other scripts in the forgery class. Also, geometrical features and a DTW-based system seem to be competitive in static signatures from Persian (Farsi) and Arabic scripts [70]. Finally, a strategy to enlarge the training set by using static synthetic signatures is tested in Indic and Western scripts in [51]. In this report on multiscript ASV, the authors used static signatures as a proof of concept.

Here again, further advances are required in online multiscript signatures, which remains an open issue.

## 6.2. Stability and Complexity

In recent years, the analysis of signature stability has been one of the key research topics in the field of signature analysis and verification. It could be said that two types of signature variability can be generally considered: while short-term variability is evident on a day-to-day basis and depends on the psychological condition of the writer and on the writing conditions, long-term variability is apparent over longer periods of time, years, or even decades and depends on the modification of the motor program in the signature writer's brain [59, 213]. According to [217], the stability can be estimated by three approaches: (1) data-based approaches use raw data to estimate signature stability, (2) feature-based approaches consider specific function features or parameters extracted. from the signature to derive stability information, and (3) model-feature approaches first describe the signatures of an individual according to a well-defined model; signature stability is then estimated by considering the range of variability of model parameters.

The stability information can be obtained by the identification of direct matching points from DTW elastic matching [206]. In this work, moreover, a multidomain signature verification is performed by using only the most stable domain of representation for each region in the signature. In other approaches, stability regions of signatures are defined as the longest similar sequences of strokes between a pair of genuine signatures [204]. This definition is based on the assumption that signing is the automated execution of a well-learned motor task and, therefore, repeated executions should ideally produce similar specimens. More recently, the analysis of stability has been applied directly to the processes underlying handwriting production [58]. For this purpose, the Sigma-Lognormal model [186] is used. The estimation of local stability is applied, starting from the extreme points of the velocity profile [210,

212].

HMMs have been considered for model-based stability analysis [97]. Stability measures have also been used to determine whether a feature contains enough information to be successfully processed by a verification system [117, 268]. When static signatures are considered, the stability of each region of a signature can be estimated by a multiple pattern matching strategy [126]. A preliminary step can be used to determine the best alignment of the corresponding regions of the signatures in order to diminish any differences between them [204]. The analysis of the optical flow is also an option for providing information about local stability [50, 209]. This technique assumes that there is a certain deformation between two genuine signatures, which is estimated on the basis of the Horn and Shunk approach. Additionally, strategies based on cosine similarity [208] or based on speeded-up robust features (SURF) [163] have been recently evaluated for estimating the stable regions in static signatures.

Signature complexity [120, 133] is another aspect that is considered with interest in the research community. In fact, for many years, the effect of signature complexity on signature verification performance has been specifically investigated [31, 69]. Notwithstanding that no common meaning of handwriting complexity has yet been defined, a signature's complexity can be thought to be an estimator of the difficulty for its imitation and therefore related to its legibility [9]. A new approach for complexity estimation is based on DMPs derived when a genuine signature is matched against a set of forgeries by DTW [211]. More recently, a statistical complexity measure, based on the concept of differential entropy in information theory, has been proposed [116]. This approach measures the complexity and stability of genuine signatures through the personal and relative entropy. A complete framework for statistical assessment of signature complexity has been also defined. This is based on static and dynamic features, for example, the number of strokes (defined in terms of lognormals in [248]), the total number of intra-/interstroke intersections, the means of intra-/interstroke intersections, the maximum of intra-/interstroke intersections, the number of x-axis intersections, the sums of the horizontal and vertical components of the velocity and acceleration at zero-crossings, and the signature length [154].

## 6.3. Acquisition Devices and Standards

Manual signature verification performed by experts has a long history [123, 214, 218]. Automatic offline and online signature verification have evolved with advancements in technology [13, 177]. For automatic offline verification, signature acquisition today is mainly performed through scanning or by photographing the signatures using digital cameras. In both cases, scanned and photographed, though the preprocessing (to deal with different distortions) differs, the underlying verification techniques are similar [160]. For online verification, signatures are usually acquired through digitized/graphic tablets, digitizing pens [158], or even gloves [135]. A graphic tablet enables a user to write his or her signatures (along with performing various other actions like drawing, writing, etc.) with the help of a pen-like device such as a stylus. These digitized signatures are then transferred to a verification engine for testing. The most common categories of graphic tablets are as follows: (1) Passive tablets: These tablets use electromagnetic induction.2 They have specialized wires spread horizontally and vertically within the tablet. These wires send and receive signals as a stylus moves in the field created by these wires (without needing the stylus to touch the tablet). The stylus draws power from the tablet beneath and does not need a battery of its own. (2) Active tablets: In such tablets, the styluses have their own batteries, giving them a heavy feel, but with less jitter as compared to passive tablets since the tablet can constantly read signals from the stylus. (3) Optical tablets: Such tablets have a special stylus with an integrated camera. The tablet underneath has an image of the dot patterns that correspond to the signature. (4) Acoustic tablets: These tablets make use of sound waves by estimating the position of a soundemitting stylus on the drawing region during the writing process. (5) Capacitive tablets: These tablets use electrostatic signals to detect the location of the stylus on the tablet, whether the stylus is touching or hovering above it.

For the above-mentioned categories of tablets, the position of the stylus, its angle of inclination on the tablet, its pressure, its proximity, and so forth are usually computed by the tablet. Apart from tablets, digital pens are also used for acquiring online signatures. A digital pen digitally captures the handwriting strokes made on paper. Digital pens differ from styluses in terms of both internal electronics and their dimensions, thickness, weight, and so forth. These pens usually have an internal memory to store information captured during the handwriting process. Most of them also possess communication channels, e.g., Bluetooth, to directly connect to computer systems. Such pens are usually shipped with software that is installed on the data-receiving computers. Once data is acquired by the pen and sent to the computer, it can be used for performing online identification or verification tasks [158]. Some important categories of digital pens are as follows: (1) inertial pens, which reconstruct trajectories, e.g., IMUPEN [262]; (2) active pens, which identify the pen location, point pressure, and so forth with the capacity to write directly onto LCDs; (3) trackball pens, which directly detect handwriting strokes written by the pen; (4) camera-based pens, which digitize handwritten strokes, e.g., the Anoto dot pattern; and (5) those that measure finger pressure during handwriting acquisition (e.g., the BiSP-M device [20]).

Since these devices are used for signatures and handwriting, there are certain standards that need to be maintained so that the accuracy and the legitimacy of the handwritten input can be ensured. These standards specify the exact requirements for the entire process that need to be fulfilled in order for the system to be considered authentic and trustworthy in the sense of being equivalent to creating an actual signature.

In the European Union, the electronic IDentification, Authentication and trust

Services (eIDAS) standard is deployed. The eIDAS focuses on the European Single Market and was formulated in 2014. The standard deals with electronic transactions and their corresponding electronic signatures. The standard, which is widely recognized, focuses on interoperability and transparency and operates across a number of platforms. Similarly, Switzerland uses the ZertES standard for electronic signature verification.

## 6.4 Signatures and Security

A well-known problem with biometrics and signatures is feature template security. In fact, solutions typically adopted with standard authentication technologies, e.g., changing the password, cannot be carried out: it is almost impossible to modify the biometric (human characteristic) trait.

Biometric feature template security relies on the use of a noninvertible feature set: the computation of the original raw data acquired by the device should be computationally difficult. In case of feature disclosure, a new set of cancellable features must be assessed. Examples of noninvertible features are total duration, number of pen-ups, sign changes in velocity and acceleration, average jerk, number of local minima, and so forth. In general, noninvertible features are parameter based. Any noninvertible transformation can be considered to be made up of several steps [157, 252]. One technique for achieving this is based on the symbolic representation of the signature's global features [107, 108]; another is histograms that represent raw feature statistics [227].

  In general, cryptography can also be considered. Cryptography refers to the study of mathematical techniques aimed at satisfying objectives related to information security. These objectives mainly concern confidentiality, integrity, authentication, and nonrepudiation. Most cryptographic techniques rely on the adoption of keys in order to encrypt and subsequently decrypt messages to be exchanged. Biometric-based authentication provides a means to not having to remember or store these keys. This is inherently a more reliable solution as biometric traits cannot be stolen or forgotten and they are extremely difficult to copy or share. One of the pioneering encrypted transaction systems was presented in [254], where authentication was complemented by additional biometric protection, based on handwritten signatures, to ensure nonrepudiation.

  A major problem here is how to unite biometrics and encryption under the requirements of confidentiality, integrity, authentication, and nonrepudiation. In fact, symmetric encryption algorithms (e.g., 3DES) fail to be compliant with the confidentiality requirement since the verification phase is not allowed in the encrypted domain. This is because the template is exposed during every verification attempt [221]. A practical solution that allows template protection is the use of asymmetric (e.g., RSA) or homomorphic encryption [100] since the verification is conducted in the encrypted domain. However, template protection has also been addressed by the "bio-cryptosystem," that is, the use of the biometric trait to manage cryptographic keys. In other words, the key is generated directly from the signature so that it is not explicitly stored [253]. The problem is that the key has to be evaluated at each verification given the signature received as input and, due to intraclass variation, it is impossible to obtain exactly the same key at each submission. In order to solve this problem, which is common to other biometrics, a fuzzy vault (FV) schema is suggested because of its ability to (fuzzily) deal with intraclass variations [132].

The FV schema has been applied to offline signature verification in [88]. The authors considered minima and maxima of the upper and lower contour of the signature as a template. They observed that the variability among users made it impossible to settle on a unique operating point of the fuzzy vault. Authors of [65] employed the concept of dissimilarity representation between the locking and the unlocking FV points. If the dissimilarity was less than a specific threshold, then the decoder succeeds in unlocking the bio-cryptographic key. Therefore, these decoders can be viewed as simple two-class thresholding classifiers that operate in a dissimilarity space. Several features aimed at enhancing accuracy and security were discussed in [67], which mainly concern adaptive matching of FV points; an ensemble of multiple FV schemes, where the key is released on the basis of a majority-vote rule; and a cascading approach, which adopts a traditional signature verification classifier before triggering the FV decoder and an adaptive key length, depending on each specific user [64]. FV has also been applied, to some extent, to online signatures [156]. The approach is based on a set of noninvertible transformations. Unfortunately, these approaches result in performance degradation along with reliability problems (i.e., the possibility of revoking a compromised template and defining a new one). Similar results have been observed when simple, noninvertible features have been used [249].

Finally, according to [94], skilled forgeries can be considered as a case of presentation attack [1]. To achieve unified criteria in biometrics and signature verification, some authors recommend introducing presentation attack detection in ASV [94].

## 6.5 Signatures for Medical Applications

For many years, the effects of health conditions on handwriting have been a field of study by graphologists and forensic experts [121, 125, 244], as well as psychologists and neuroscientists [153]. Handwriting is a complex activity resulting from cognitive, kinesthetic, and perceptual motor abilities [185, 231]. Furthermore, visual and kinesthetic perception, motor planning, eye-hand coordination, visual-motor integration, and manual skills are involved in this sensorimotor task. For this reason, psychiatric, neurological, or peripheral illnesses can generally have strong effects on handwriting, so changes in handwriting are good candidates as biomarkers for the assessment of several diseases. As the number of devices for data capturing and processing is increasing, the use of handwriting to detect and monitor health conditions is becoming more and more attractive. Handwriting analysis will not, of course, compete with traditional medical strategies. It can provide a different and complementary approach to analyzing health problems in a noninvasive way and at low cost .

Recent research has focused on relations between handwriting and Parkinson's disease [225], Alzheimer's disease [35], agraphia [237], dysgraphia [41], psychosis [34], and other pathologies [33, 60, 168]. For instance, the handwriting of individuals affected by Parkinson's disease [38, 231] is smaller than normal; moreover, it is produced more slowly and with lower pressure on the writing tablet [224, 239]. Conversely, although it has been observed that handwriting impairment is heterogeneous within the population affected by Alzheimer's disease, there are certain aspects of the writing process that are more vulnerable than others and may present diagnostic signs. In particular, handwriting of individuals affected by Alzheimer's disease is less consistent in length and duration [269]. In general, it has been found that during the clinical course, dysgraphia occurs during both the early and the later stages of the disorder progression [238, 270].

In light of the observations above, handwriting could be considered as a valuable tool for characterizing aging [265, 266] and for detecting early signs of disease [113, 187–189, 215, 266], as well as for monitoring the evolution of illnesses [32] and the effect of therapies [24], and it can be used as a tool for the analysis of the emotional state [68]. Despite the large number of works focusing their attention on handwriting tasks in general, few studies have focused specifically on handwritten signatures. In the following, some of the most recent work is taken into account. In [155], the relation between handwritten signatures and personality traits was studied by considering both the static and dynamic features of signatures. The results reveal significant correlations between personality traits and these features. In addition, it was found that gender and weight could be predicted effectively, especially using signature velocity characteristics. Conversely, the relation between the handwritten signature, and more precisely the signature position on or below a dotted line, and other cognitive functions was further investigated in [261]. The findings of the study suggest that signature positioning may be a marker of vulnerability for the mechanisms involved in visuospatial abilities.

A machine-learning algorithm to automatically discriminate between Alzheimer's disease patients and healthy control subjects was employed in [207]. The work was based on a previously annotated dataset of handwritten signatures. For this purpose, the Sigma-Lognormal model [186] was used to extract dynamic features of the signatures and to obtain sets of attributes for building classification models. Experimental results show that the proposed approach is inexpensive and effective. Therefore, it can be considered as a promising direction for future research, for example, for implementing a screening standard routine for the early diagnosis of Alzheimer's disease. The work published in [222] was specifically focused on initial cognitive impairment and investigated whether it is possible to make inferences on decision-making capacity through the analysis of signatures and spontaneous writing. Unfortunately, while significant correlations were found between spontaneous writing indices and neuropsychological test results, no correlation was found between signature deterioration and cognitive decline. Therefore, the authors concluded that the integrity of the graphic pattern of a signature is not indicative of the cognitive capacity of individuals with cognitive impairment, so they recommend using great caution in attributing validity to the signature of such individuals for this purpose.

The objective in [224] was aimed at finding simple characteristics of handwriting capable of accurately differentiating between Parkinson's disease patients and healthy control subjects. Participants in this study were requested to write their name and to copy an address. Statistical analysis, used to test group differences in spatial, temporal, and pressure measures for each of the writing tasks, showed significant group effects: compared to the control, patients wrote smaller letters, applied less pressure, and required more performance time. In conclusion, the authors highlighted the importance of analyzing handwriting not only "on paper" but also "in air," i.e., when the pen has not yet touched the paper, since significant differences were observed between these two writing conditions. In fact, "in air" time is a manifestation of the "planning of the next movement," which reflects the cognitive ability and supplies information about the writer [233]. A similar study was conducted in [224], in which the analysis of name writing, together with the analysis of two other tasks, i.e., paragraph copying and freestyle writing, was applied to the study of the characteristics of children affected by high-functioning autism spectrum disorder. The handwriting process analysis across the three tasks indicates that the pen tilt of the children affected by the disorder was significantly lower than control subjects. Their pen stroke duration "on paper" was significantly higher, as were their letter sizes. Furthermore, their stroke duration "in air" was significantly higher in the name and freestyle writing tasks. It is worth noting the works in [224, 226] that focus on the analysis of name writing, which is a slightly different task from signing.

## 6.6. Synthetic Signature Generation Models

Models and methods for generating synthetic handwritten signatures provide an opportunity to clarify and learn the biological processes that characterize such specimens. Among the number of advantages of synthesized signatures, some of the most relevant can be summarized as follows: (1) it is relatively effortless to produce a large amount of data (once the generation algorithm has been developed); (2) there is no size restriction (in terms of subjects and samples per subject) since the signatures are automatically produced by a computer program; (3) such signatures are not subject to legal constraints because they do not relate to a real user; (4) they eliminate human mistakes such as labeling the data, which biases the performance evaluation of the algorithms; (5) they permit statistically meaningful evaluations of performance; (6) they can simulate aging or maturity level models; (7) they can simulate signatures affected by some behavioral disorders, neurodegenerative diseases, or other cognitive impairments; and (8) they can provide an opportunity to analyze the deterioration and loss of function in the organs involved in handwriting production.

The literature contains a number of novel proposals for the generation of synthetic identities. In some, the algorithms define a new identity at the beginning without any real signature as a reference. Such models are then able to generate realistic intrapersonal variability of a virtual writer. In this context, the statistical normality of signatures' lexical morphology [55] is useful for such purposes. One of the uses of these synthetic signatures is to test the systems in order to create common benchmarks with meaningful statistical results. Additionally, signature generation is dependent on the script that has to be reproduced. Thus, special attention to the peculiarities in a given script needs to be taken into account. For instance, Western signatures are typically composed of text and flourishes, Chinese or Japanese signatures are designed with symbols, and so on. The full generation of online, flourish-based signatures using two algorithms was carried out in a linked pair of papers [93, 96]. One of the strategies was based on spectral analysis and the other on the kinematic theory of rapid human movement. Apart from a visual validation, quantitative evaluations were carried out, mainly in terms of comparative performance with real databases. On the generation of offline signatures, a method based on the statistical distribution of global signature properties to generate flourish-based Western specimens can be found in [76]. In their proposal, the authors generated both genuine signatures and forgeries. A more elaborate procedure to generate signatures based on equivalence theory was proposed for Western offline signatures [73] and for Western online signatures [75]. These authors defined a user grid based on the cognitive map in order to represent the name and the flourish engrams. The dynamics of the signature trajectory are calculated by lognormal sampling of the signature trajectory. The static version is eventually generated by applying an ink deposition model. A similar proposal for generating offline Bengali and Devanagari signatures is reported in [47, 74].

Many proposals focus on modeling intrapersonal variability for duplicating static or dynamic signatures (e.g., [70, 77, 90, 92, 106]). In this context, duplicating a signature means generating a new specimen artificially from one or more real signatures. Among its advantages, signature duplication can improve the training of ASV systems and allows the carrying out of statistically meaningful evaluations as well as enlarging the number of signatures in databases. It also helps match the baseline performances for real signatures, which are often difficult to obtain, and it can improve the performance of existing automatic signature verifiers. The dynamic duplicated signature performance can be comparable to that for real signatures. By applying a random and new deformation, it is possible to improve the performance of an HMM-based classifier [92]. To increase the training set, a method based on a clonal selection of the enrolled signatures is studied in [242]. The properties of the kinematic theory of rapid movement have been used for these purposes [52, 53]. Additionally, other authors have worked on the replication of multiple offline specimens from real online signatures applying geometrical [119] or cognitive based [54] strategies. Other proposals have focused on duplicating offline signatures from offline specimens. In [90], an offline signature dataset is enlarged by applying some geometrical transformations. Another proposal applies neuroscience perspectives in a duplication method [48].

This work on synthetic signatures opens the opportunity to explore more intelligent systems as well as to understand better interrelated information, which depends on the nature of the signatures. Indeed, the potential relationship between online and offline signatures is still a matter of research. An ink deposition model is elaborated in [77] to generate static signatures from online ones. Some strategies based on the modification of online sampling rates and image resolution were proposed in [106]. The dynamic information (i.e., velocity, pressure) was used in [56, 91] to create images based on overlapping 2D Gaussian spots. Recovering the dynamic information from a static signature to the best of our knowledge, is a topic that has not yet been examined in any depth [49, 180], although many attempts have been made to address it for handwriting, e.g., in [245]. The most critical stages relate to thinning and writing order recovery [180].

## 6.7 Signing on Touchscreens

The recent integration of low-cost data acquisition devices in a multitude of personal systems, i.e. smartphones, PDAs, and so forth, has stimulated several researchers into investigating dynamic signature verification on mobile devices [227]. Clearly, signature verification on such devices is affected by several factors not present in traditional devices specifically conceived for handwriting recognition. Mobile devices, in fact, are characterized by a small input area and poor ergonomics, which affect user interaction and lead to large intraclass variability [13]. The quality of signature acquisitions can show high disparity because of the quality of the touchscreens and because the amount of information that can be captured is limited as pressure, pen-azimuth, and other attributes, which could lead to improved performance, cannot typically be recorded. Furthermore, users are asked to sign on an unstable surface, often using a single finger [14, 177].

Migrating to mobile devices from traditional pen tablets has therefore raised several new issues: fine-tuning the verification algorithm for improving the performance, considering the execution platform and the acquisition properties (e.g., [25, 26, 118, 119, 143, 166–168, 172, 251, 273]); evaluating the impact on the performance of the implementation of the algorithm in a variety of platforms [249]; evaluating the usability of the new applications and redesigning them in accordance with the results obtained [27]; and guaranteeing the security of the biometric data in the new mobile context [249].

Concerning performance, the article published in [166] was among the first to recognize the importance of adapting the traditional tablet-based signature verification systems in the context of handheld devices. This work presented a systematic comparison between the system proposed and a traditional pen-tablet-based system by employing a subset of 120 users from BioSecure DS2 and DS3. BioSecure DS2 contains data captured under a control scenario where users had to sign while sitting; BioSecure DS3 considers a mobile scenario where users had to sign while standing and holding a PDA in one hand. The low discrimination power of dynamic features (time, speed, and acceleration) observed in these PDA scenarios suggested that ergonomics, an unfamiliar surface, and the signing device (touchscreen and PDA stylus vs. traditional pen and paper) affect the dynamics of the signature acquisition process. In [171], the possibility of applying a pre-existing online signature verification algorithm to portable devices was studied. To gain insight into the performance of the algorithm prior to its real implementation, the authors modified the MCYT signature database to simulate signatures taken from a mobile device. In particular, two main screen technologies were simulated: capacitive and resistive screens [141]. As expected by the authors, simulations of the resistive screens yielded better results than those obtained by simulating capacitive screens, as the former provide pressure information. In [167], the discriminative power of global and local features between mobile devices and pen tablets was compared. Results showed a decrease in the feature discriminative power and a higher verification error rate on handheld devices. It was found that one of the main causes of performance degradation on handheld devices was the absence of pen-up trajectory information. A mobile-biometric signature verification system was proposed in [273], which was based on ANNs and a comparative analysis of the performance of the proposed system for two datasets: one obtained using a standard device for biometric signatures, the other a mobile database obtained using a smartphone. The experimental results showed that the performance of mobile devices was low when compared to traditional biometric signature capturing devices. The main reason the authors found for this was the absence of pen-tilt angle information in the mobile device dataset. Nevertheless, the variability in these databases is also significantly greater as the users are accustomed to signing on traditional tablets. Due to the large variety of electronic devices employed today for signature acquisition, another relevant direction of research concerns the problem of interoperability. To investigate this problem, a two-stage approach was used in [250]. The former is a preprocessing stage where data acquired from different devices are preprocessed to normalize the signals in similar ranges. The latter consists of selecting the features that are most robust for further reducing the effect of device interoperability.

Usability is also a field of active research. The end-user who interacts with the product can be dissatisfied with it. This can involve misuse because of poor results or even complete rejection of the technology. In light of these observations, three experiments aimed at obtaining usability conclusions for improving future biometric implementations were carried out in [27]. The experiments considered different devices, platforms, and technologies. Various scenarios, in which users had to sign in different postures, were analyzed. The authors observed that the stylus-based devices obtained better results in performance when the user was sitting on a chair and the device was resting on a table (the most common situation for signing). On the other hand, the finger-based devices returned the best results in performance within the scenarios where the user had to handle the device without support. The better the visual feedback, the better are the performance and the usability: in fact, it was observed that users did not feel comfortable when no visual feedback was provided. Stress influenced negatively both performance and usability. Nevertheless, probably due to the increasing stress at the moment of signing, participants became used to signing quickly and carelessly, so that the performance results were close to the ones obtained without stress.

Another important research direction concerns the security of our sensitive data in the context of mobile devices. Due to the high deployment of mobile devices and their increasing popularity in our society, the use of biometric traits, for example, in commercial and banking applications, provides an easy, quick, and reliable way to perform payments through these devices. As was remarked in [249], most state-of-the-art systems are based on extracting information contained in the trajectory coordinates of the signing process, which is stored in a biometric template. Protecting this sensitive information against possible external attack is critical: such attacks, in fact, would allow criminals to perform direct attacks on a biometric system or carry out high-quality forgeries of the users' signatures. To this end, in [249], the extreme case of not considering the information related to the trajectory and its derivatives in a biometric system was also studied. Such a system would be robust against attacks, as the critical biometric information would not be stored anywhere.

At present, the more common dynamic signature verification systems adopt the contact mode of signing on the mobile device. Therefore, these systems have a great dependence on the device technology. Considering that the development of biometrics seems to tend toward the long-distance and noncontact modalities, in-air signature verification systems [18, 71, 105] open novel perspectives that demand further investigation.

It is important to note that a large database tailored to study signature verification in mobile scenarios, namely, e-BioSign, has recently been made available to the research community [246]. This database was designed to collect data from five devices: three of them were Wacom devices specifically developed for signature and handwriting applications; the other two were general purpose Samsung tablets that could collect data not only by using a pen stylus but also by using a finger. The e-BioSign database may well enable future research toward a better understanding of human handwriting when captured in mobile scenarios.

It must be stated that, up to now, the same feature set has been considered whenever the signature has been written by using a finger or a stylus. Occasionally, some features have been removed in order to allow comparison or for security purposes. However, in general, the features that are considered are those related to the stylus domain. The finger-signing scenario allows the consideration of features from the keystroke and touch dynamic domains [174, 228]. The touch size and the shape of the contact area have apparently never been considered within the task of signature verification. These features could be coupled with standard features and could have an interesting impact since those features can be considered, to some extent, to belong to the set of physiological biometrics (e.g., hand geometry, finger size, etc.). Moreover, the future availability of under display fingerprint scanners will allow the synchronous combination of fingerprint and signature data [114]. Signing with a finger over a display also allows the taking into account of tapping behavior [276]. In fact, it has been shown that touch analysis can be used to recognize users [87, 179].

## 7. CONCLUSION

We have presented in this article an update on automatic signature verification and a prospective analysis of an active and dynamic field of research that challenges the image and signal processing research communities.

An analysis has been carried out of the weaknesses and strengths of the most common publicly available offline and online signature databases. ASV systems have traditionally been developed to cope with Western-based signatures. However, this article also pays attention to the growing use of other scripts, e.g., Arab and Persian, among others. Beyond being a sign of maturity, such technological growth can be attributed to the high acceptability, in many cultures and countries, of the handwritten signature for use in authentication. Nevertheless, a comparison among the proposed ASV systems is really difficult to carry out. One of the main constraints is the different evaluation protocols in the different proposals. However, different competitions in handwritten signatures have been organized in the last decade, where it is clearly possible to assess the progress of this technology. Thus, results from the competitions can be seen as the current state of the art in this field, since they can guarantee the same benchmark under the same conditions. One of the most important areas seen in the last 10 years is the tendency to design systems useful for forensic handwriting experts. To this end, this article also reviews new proposals in this field. Additionally, some novelties and progress can be detected in the perspective analysis carried out here. We also include a summary of the uses of handwritten signatures in other related fields, which are briefly listed as developments on stability and complexity measurements on signatures, international standards, advances in cryptosystems, the use of signatures on e-health applications, and the generation of models to synthesize handwritten signatures.

On top of reporting the most recent results, we have pointed out the new trends that might constitute the driving force for the next breakthrough in this field. To conclude, we can look into our crystal ball and go even further in these directions. Among these tendencies, it becomes more and more clear that the offline applications, on the one hand, will align more and more with forensic practice, but, on the other hand, the main offline deployments will have to come back to large public applications. Such key applications are far from being evident with the decrease of check payments and paper contract signing, so this will probably rely more on the need for multiscript personalized cryptography, where the number of practical uses is booming. In this context, Sigma Cryptography is expected to become a key player, combining the naturalness of identifying ourselves by actual signatures with the robustness of digital signatures. From the online domain, the widespread availability of handheld devices will become the driving force. Among other things, these mobile units will allow in the near future the merging of the handwriting and signing interactions into what has been recently labeled as personal digital bodyguards (PDBs) [216]. PDBs will be able to supplement people's sensitive data protection with signature verification. Such equipment will use writer authentication and personalized handwritten CAPTCHAs processing and enhance human-machine interaction performance through word spotting and handwriting recognition. For young children, these tools will turn into interactive tools helping them to learn and master their fine motor control and become better at handwriting. PDBs will also be able to provide the user with fine motor control monitoring, which can detect stress, aging, and health problems, thus bringing medical applications to the forefront.

Finally, as for many fields where successful commercial applications have been developed, these breakthroughs come from the development of robust algorithms tested and validated on huge representative databases, from which benchmarks can be designed and comparative analysis can be conducted. There is a move in this direction as we have seen using both real and synthesized signatures. This will become a must in the next decade if we expect to design robust signature verification systems based on a small number of references at registration. These systems need to evolve with time and reach and maintain high accuracy.

## REFERENCES

[1] ISO/IEC 30107-1:2016. 2016. Information technology - Biometric presentation attack detection - Part 1: Framework. Int. Org. Standardization, Geneva.

[2] I. A. B. Abdelghani and N. E. B. Amara. 2013. SID signature database: A Tunisian off-line handwritten signature database. In Int. Conf. on Image Analysis and Processing. Springer, 131–139.

[3] S. M. S. Ahmad, A. Shakil, A. R. Ahmad, M. A. M. Balbed, and R. M. Anwar. 2008. SIGMA-A Malaysian signatures' database. In IEEE/ACS Int. Conf. on Computer Systems and Applications. IEEE, 919–920.

[4] K. Ahrabian and B. Babaali. 2017. On usage of autoencoders and siamese

networks for online handwritten signature verification. Corr (2017). arXiv:1712.02781

[5] Y. Akbari, M. J. Jalili, J. Sadri, K. Nouri, I. Siddiqi, and C. Djeddi. 2018. A novel database for automatic processing of Persian handwritten bank checks. Pattern Recognition 74 (2018), 253–265.

[6] S. Al-Maadeed, W. Ayouby, A. Hassaine, A. Almejali, A. Al-Yazeedi, and R. Al-Atiya. 2012. Arabic signature verification dataset. In 13th Int. Arab Conf. on Information Technology. 288–293.

[7] A. Alaei, S. Pal, U. Pal, and M. Blumenstein. 2017. An efficient signature verification method based on an interval symbolic representation and a fuzzy similarity measure. IEEE Transactions on Information Forensics and Security 12, 10 (2017), 2360–2372.

[8] L. C. Alewijnse, C. E. van den Heuvel, R. D. Stoel, and K. Franke. 2009. Analysis of signature complexity. In Proceedings of the 14th Biennial Conf. of the Int. Graphonomics Society: Advances in Graphonomics. 6–9.

[9] F. Alonso-Fernandez, M. C. Fairhurst, J. Fierrez, and J. Ortega-Garcia. 2007. Impact of signature legibility and signature type in off-line signature verification. In Biometrics Symposium, 2007. IEEE, 1–6.

[10] F. Alonso-Fernandez, J. Fierrez, A. Gilperez, J. Galbally, and J. Ortega-Garcia. 2009. Robustness of signature verification systems to imitators with increasing skills. In 10th Int. Conf. on Document Analysis and Recognition. IEEE, 728–732.

[11] G. Alvarez, B. Sheffer, and M. Bryant. 2016. Offline Signature Verification with Convolutional Neural Networks. Technical Report. Stanford University, Stanford. [12] A. Q. Ansari, M. Hanmandlu, J. Kour, and A. K. Singh. 2013. Online signature verification using segment-level fuzzy modelling. IET Biometrics 3, 3 (2013), 113–127.

[13] M. Antal and A. Bandi. 2017. Finger or stylus: Their impact on the performance of on-line signature verification systems. In Proceedings of the 5th International Conference on Recent Achievements in Mechatronics, Automation, Computer Sciences and Robotics (MACRo'17).

[14] M. Antal, L. Z. Szabó, and T. Tordai. 2018. Online signature verification on MOBISIG finger-drawn signature corpus. Mobile Information Systems 2018, Article 3127042 (2018) 15 pages. https://doi.org/10.1155/2018/3127042.

[15] E. Argones Rua and J. L. Alba Castro. 2012. Online signature verification based on generative models. IEEE Transactions on Systems, Man, and Cybernetics Part B (Cybernetics) 42, 4 (2012), 1231–1242.

[16] E. Argones Rua, E. Maiorana, J. L. Alba Castro, and P. Campisi. 2012. Biometric template protection using universal background models: An application to online signature. IEEE Transactions on Information Forensics and Security 7, 1 (2012), 269–282.

[17] C. E. H. Berger, B. Robertson, and G. A. Vignaux. 2016. Interpreting Evidence: Evaluating Forensic Science in the Courtroom. Vol. 2. Wiley.

[18] G. Bailador, C. Sanchez-Avila, J. Guerra-Casanova, and A. de Santos Sierra. 2011. Analysis of pattern recognition techniques for in-air signature biometrics. Pattern Recognition 44, 10 (2011), 2468–2478.

[19] K. Barkoula, G. Economou, and S. Fotopoulos. 2013. Online signature verification based on signatures turning angle representation using longest common subsequence matching. International Journal on Document Analysis and Recognition 16, 3 (2013), 261–272.

[20] M. Bashir and F. Kempf. 2012. Advanced biometric pen system for recording and analyzing handwriting. Journal of Signal Processing Systems 68, 1 (2012), 75–81.

[21] L. Batista, E. Granger, and R. Sabourin. 2012. Dynamic selection of generative–discriminative ensembles for off-line signature verification. Pattern Recognition 45, 4 (2012), 1326–1340.

[22] D. Bertolini, L. S. Oliveira, E. Justino, and R. Sabourin. 2010. Reducing forgeries in writer-independent off-line signature verification through ensemble of classifiers. Pattern Recognition 43, 1 (January 2010), 387–396.

[23] B. Bhanu and A. Kumar. 2017. Deep Learning for Biometrics. Springer.

[24] C. Bidet-Ildei, P. Pollak, S. Kandel, V. Fraix, and J.-P. Orliaguet. 2011. Handwriting in patients with Parkinson disease: Effect of L-dopa and stimulation of the sub-thalamic nucleus on motor anticipation. Human Movement Science 30, 4 (2011), 783–791. ACM Computing Surveys, Vol. 51, No. 6, Article 117. Publication date: January 2019. 117:30 M. Diaz et al.

[25] R. Blanco-Gonzalo, O. M. Hurtado, A. Mendaza-Ormaza, and R. Sanchez-Reillo. 2012. Handwritten signature recognition in mobile scenarios: Performance evaluation. In IEEE Int. Carnahan Conf. on Security Technology (ICCST'12). 174–179.

[26] R. Blanco-Gonzalo, R. Sanchez-Reillo, O. M. Hurtado, and J. Liu-Jimenez. 2013. Performance evaluation of handwritten signature recognition in mobile environments. IET Biometrics 3, 3 (2013), 139–146.

[27] R. Blanco-Gonzalo, R. Sanchez-Reillo, O. M. Hurtado, and J. Liu-Jimenez. 2013. Usability analysis of dynamic signature verification in mobile environments. In Int. Conf. of the Biometrics Special Interest Group (BIOSIG'13). IEEE, 1–9.

[28] V. L. Blankers, C. E. v. d. Heuvel, K. Y. Franke, and L. G. Vuurpijl. 2009. ICDAR 2009 signature verification competition. In 10th Int. Conf. on Document Analysis and Recognition (ICDAR'09). 1403–1407.

[29] M. Blumenstein, M. A. Ferrer, and J. F. Vargas. 2010. The 4NSigComp2010 off-line signature verification competition: Scenario 2. In Int. Conf. on Frontiers in Handwriting Recognition, (ICFHR'10). 721–726.

[30] W. Bouamra, C. Djeddi, B. Nini, M. Diaz, and I. Siddiqi. 2018. Towards the design of an offline signature verifier based on a small number of genuine samples for training. Expert Systems with Applications 107 (2018), 182–195.

[31] J. J. Brault and R. Plamondon. 1993. A complexity measure of handwritten curves: Modeling of dynamic signature forgery. IEEE Transactions on Systems, Man, and Cybernetics 23, 2 (1993), 400–413.

[32] M. P. Caligiuri and L. A Mohammed. 2012. The Neuroscience of Handwriting: Applications for Forensic Document Examination. CRC Press.

[33] M. P. Caligiuri, H.-L. Teulings, C. E. Dean, and J. B. Lohr. 2015. A quantitative measure of handwriting dysfluency for assessing tardive dyskinesia. Journal of

Clinical Psychopharmacology 35, 2 (2015), 168.

[34] M. P. Caligiuri, H.-L. Teulings, C. E. Dean, A. B Niculescu, and J. B. Lohr. 2010. Handwriting movement kinematics for quantifying extrapyramidal side effects in patients treated with atypical antipsychotics. Psychiatry Research 177, 1 (2010), 77–83.

[35] A. Carina-Fernandes and J. M. Lopes-Lima. 2017. Alzheimer's disease and handwriting. what do we know so far? In 18th Biennial Conf. of the Int. Graphonomics Society. 131–134.

[36] G. Chassang. 2017. The impact of the EU general data protection regulation on scientific research. Ecancermedicalscience 11, 709 (2017). DOI:10.3332/ecancer.2017.709

[37] Z. Chen, X. Xia, and F. Luan. 2016. Automatic online signature verification based on dynamic function features. In 7th IEEE Int. Conf. on Software Engineering and Service Science (ICSESS'16). 964–968.

[38] F. Costa and A. Accardo. 2017. Kinematic analysis of handwriting in parkinson disease. In 18th Conf. of the Int. Graphonomics Society (IGS'17). 135–138.

[39] K. Cpałka and M. Zalasiński. 2014. On-line signature verification using vertical signature partitioning. Expert Systems with Applications 41, 9 (2014), 4170–4180.

[40] K. Cpałka, M. Zalasiński, and L. Rutkowski. 2014. New method for the on-line signature verification based on horizontal partitioning. Pattern Recognition 47, 8 (2014), 2652–2661.

[41] J. Danna, V. Paz-Villagrán, C. Gondre, M. Aramaki, R. Kronland-Martinet, S. Ystad, and J.-L. Velay. 2013. Handwriting sonification for the diagnosis of dysgraphia. In Recent Progress in Graphonomics: Learn from the Past–Proceedings of the 16th Conf. of the Int. Graphonomics Society. 123–126.

[42] S. Darwish and A. El-Nour. 2016. Automated offline arabic signature verification system using multiple features fusion for forensic applications. Arab Journal of Forensic Sciences & Forensic Medicine 1, 4 (2016), 424–437.

[43] A. Das, M. A Ferrer, U. Pal, S. Pal, M. Diaz, and M. Blumenstein. 2016. Multi-script versus single-script scenarios in automatic off-line signature verification. IET Biometrics 5, 4 (2016), 305–313.

[44] E. Dehghani and M. E. Moghaddam. 2009. On-line signature verification using ANFIS. In 6th Int. Symposium on Image and Signal Processing and Analysis. 546–549.

[45] H.-R. Deng and Y.-H. Wang. 2009. On-line signature verification based on correlation image. In Int. Conf. on Machine Learning and Cybernetics, Vol. 3. 1788–1792.

[46] S. Dey, A. Dutta, J. I. Toledo, S. K. Ghosh, J. Lladós, and U. Pal. 2017. SigNet: Convolutional Siamese network for writer independent offline signature verification. Corr (2017). arXiv:1707.02131

[47] M. Diaz, S. Chanda, M. A. Ferrer, C. K. Banerjee, A. Majumdar, C. Carmona-Duarte, P. Acharya, and U. Pal. 2016. Multiple generation of bengali static signatures. In 15th Int. Conf. on Frontiers in Handwriting Recognition (ICFHR'16). 42–47.

[48] M. Diaz, M. A Ferrer, G. S Eskander, and R. Sabourin. 2017. Generation of

duplicated off-line signature images for verification systems. IEEE Transactions on Pattern Analysis and Machine Intelligence 39, 5 (2017), 951–964.

[49] M. Diaz, M. A. Ferrer, A. Parziale, and A. Marcelli. 2017. Recovering western on-line signatures from image-based specimens. In 14th IAPR Int. Conf. on Document Analysis and Recognition (ICDAR'17). 1204–1209.

[50] M. Diaz, M. A. Ferrer, G. Pirlo, G. Giannico, and D. Impedovo. 2015. Off-line signature stability by optical flow: Feasibility study of predicting the verifier performance. In 49th IEEE Int. Carnahan Conf. on Security Technology (ICCST'15). 341–345. ACM Computing Surveys, Vol. 51, No. 6, Article 117. Publication date: January 2019. A Perspective Analysis of Handwritten Signature Technology 117:31

[51] M. Diaz, M. A. Ferrer, and R. Sabourin. 2016. Approaching the intra-class variability in multi-script static signature evaluation. In 23rd Int. Conf. on Pattern Recognition (ICPR'16). IEEE, 1147–1152.

[52] M. Diaz, A. Fischer, M. A. Ferrer, and R. Plamondon. 2018. Dynamic signature verification system based on one real signature. IEEE Transactions on Cybernetics 48, 1 (2018), 228–239.

[53] M. Diaz, A. Fischer, R. Plamondon, and M. A. Ferrer. 2015. Towards an automatic on-line signature verifier using only one reference per signer. In Proc. IAPR Int. Conf. on Document Analysis and Recognition (ICDAR'15). 631–635.

[54] M. Diaz-Cabrera, M. A. Ferrer, and A. Morales. 2014. Cognitive inspired model to generate duplicated static signature images. In Int. Conf. on Frontiers in Handwriting Recognition (ICFHR'14). 61–66.

[55] M. Diaz-Cabrera, M. A. Ferrer, and A. Morales. 2015. Modeling the lexical morphology of western handwritten signatures. PLoS ONE 10, 4 (2015), e0123254. https://journals.plos.org/plosone/article/citation?id=10.1371/journal. pone.0123254.

[56] M. Diaz-Cabrera, M. Gomez-Barrero, A. Morales, M. A. Ferrer, and J. Galbally. 2014. Generation of enhanced synthetic off-line signatures based on real on-line data. In 14th Int. Conf. on Frontiers in Handwriting Recognition (ICFHR'14). 482–487.

[57] M. Diaz-Cabrera, A. Morales, and M. A Ferrer. 2014. Emerging issues for static handwritten signature biometric. In Advances in Digital Handwritten Signature Processing. A Human Artefact for e-Society. 111–122.

[58] M. Djioua and R. Plamondon. 2009. A new algorithm and system for the characterization of handwriting strokes with delta-lognormal parameters. IEEE Transactions on Pattern Analysis and Machine Intelligence 31, 11 (2009), 2060–2072.

[59] M. Djioua and R. Plamondon. 2009. Studying the variability of handwriting patterns using the kinematic theory. Human Movement Science 28, 5 (2009), 588–601.

[60] T. A. Duda, J. E. Casey, and N. McNevin. 2014. Variability of kinematic graphomotor fluency in adults with ADHD. Human Movement Science 38 (2014), 331–342.

[61] A. Dutta, U. Pal, and J. Lladós. 2016. Compact correlated features for writer independent signature verification. In 23rd Int. Conf. on Pattern Recognition (ICPR'16). IEEE, 3422–3427.

[62] H. Edwards, C. Gotsonis, et al. 2009. Strengthening Forensic Science in the

United States: A Path Forward. National Academies Press.

[63] Y. Elmir, Z. Elberrichi, and R. Adjoudj. 2014. Multimodal biometric using a hierarchical fusion of a person's face, voice, and online signature.Journal of Information Processing Systems 10, 4 (2014), 555–567.

[64] G. S Eskander, R. Sabourin, and E. Granger. 2014. A bio-cryptographic system based on offline signature images. Information Sciences 259 (2014), 170–191.

[65] G. S. Eskander, R. Sabourin, and E. Granger. 2011. Signature based fuzzy vaults with boosted feature selection. In Workshop on Computational Intelligence in Biometrics and Identity Management (CIBIM'11). IEEE, 131–138.

[66] G. S. Eskander, R. Sabourin, and E. Granger. 2013. Hybrid writer-independent–writer-dependent offline signature verification system. IET Biometrics 2, 4 (2013), 169–181.

[67] G. S. Eskander, R. Sabourin, and E. Granger. 2014. Offline signature-based fuzzy vault: A review and new results. In Symposium on Computational Intelligence in Biometrics and Identity Management. 45–52.

[68] M. Fairhurst, M. Erbilek, and C. Li. 2014. Enhancing the forensic value of handwriting using emotion prediction. In Int. Workshop on Biometrics and Forensics (IWBF'14). 1–6.

[69] M. C. Fairhurst, E. Kaplani, and R. M. Guest. 2001. Complexity measures in handwritten signature verification. In Proc. 1st Intl. Conf. on Universal Access in Human-Computer Interaction.

[70] D. Falahati, M. S. Helforoush, H. Danyali, and M. Rashidpour. 2011. Static signature verification for Farsi and Arabic signatures using dynamic time warping. In 19th Iranian Conf. on Electrical Engineering (ICEE'11). 1–6.

[71] Y. Fang, W. Kang, Q. Wu, and L. Tang. 2017. A novel video-based system for in-air signature verification. Computers & Electrical Engineering 57 (2017), 1–14.

[72] S. A. Farimani and M. V. Jahan. 2018. An HMM for online signature verification based on velocity and hand movement directions. In 6th Iranian Joint Congress on Fuzzy and Intelligent Systems. 205–209.

[73] M. A. Ferrer, M. Diaz-Cabrera, and A. Morales. 2015. Static signature synthesis: A neuromotor inspired approach for biometrics. IEEE Transactions on Pattern Analysis and Machine Intelligence 37, 3 (2015), 667–680.

[74] M. A. Ferrer, S. Chanda, M. Diaz, C. K. Banerjee, A. Majumdar, C. Carmona-Duarte, P. Acharya, and U. Pal. 2018. Static and dynamic synthesis of Bengali and Devanagari signatures. IEEE Transactions on Cybernetics 48, 10 (2018), 2896–2907.

[75] M. A Ferrer, M. Diaz, C. Carmona-Duarte, and A. Morales. 2017. A behavioral handwriting model for static and dynamic signature synthesis. IEEE Transactions on Pattern Analysis and Machine Intelligence 39, 6 (2017), 1041–1053.

[76] M. A Ferrer, M. Diaz-Cabrera, and A. Morales. 2013. Synthetic off-line signature image generation. In Int. Conf. on Biometrics. IEEE, 1–7.

[77] M. A. Ferrer, M. Diaz-Cabrera, A. Morales, J. Galbally, and M. Gomez-Barrero. 2013. Realistic synthetic off-line signature generation based on synthetic on-line data. In Proc. IEEE Int. Carnahan Conf. on Security Technology (ICCST'13). 116–121. ACM Computing Surveys, Vol. 51, No. 6, Article 117. Publication date: January 2019. 117:32 M. Diaz et al. [78] M. A Ferrer, J. F. Vargas, A. Morales, and

A. Ordóñez. 2012. Robustness of offline signature verification based on gray level features. IEEE Transactions on Information Forensics and Security 7, 3 (2012), 966–977.

[79] J. Fierrez, J. Galbally, J. Ortega-Garcia, M. R. Freire, F. Alonso-Fernandez, D. Ramos, D. T. Toledano, J. GonzalezRodriguez, J. A. Siguenza, J. Garrido-Salas, E. Anguiano, G. Gonzalez-de-Rivera, R. Ribalda, M. Faundez-Zanuy, J. A. Ortega, V. Cardenoso-Payo, A. Viloria, C. E. Vivaracho, Q. I. Moro, J. J. Igarza, J. Sanchez, I. Hernaez, C. OrriteUruñuela, F. Martinez-Contreras, and J. J. Gracia-Roche. 2010. BiosecurID: A multimodal biometric database. Pattern Analysis and Applications 13, 2 (2010), 235–246.

[80] J. Fierez and J. Ortega-Garcia. 2008. On-line Signature Verification. Springer, 189–209.

[81] J. Fierrez-Aguilar, N. Alonso-Hermira, G. Moreno-Marquez, and J. Ortega-Garcia. 2004. An off-line signature verification system based on fusion of local and global information. In International Workshop on Biometric Authentication. Springer, 295–306.

[82] A. Fischer, M. Diaz, R. Plamondon, and M. A. Ferrer. 2015. Robust score normalization for DTW-based on-line signature verification. In 13th Int. Conf. on Document Analysis and Recognition (ICDAR'15). 241–245.

[83] A. Fischer and R. Plamondon. 2017. Signature verification based on the kinematic theory of rapid human movements. IEEE Transactions on Human-Machine Systems 47, 2 (2017), 169–180.

[84] A. Foroozandeh, Y. Akbari, M. J. Jalili, and J. Sadri. 2012. Persian signature verification based on fractal dimension using testing hypothesis. In Int. Conf. on Frontiers in Handwriting Recognition (ICFHR'12). 313–318.

[85] B. Found. 2011. An introduction to the character of forensic investigations into questioned signatures. In 1st Int. Workshop on Automated Forensic Handwriting Analysis (AFHA'11).

[86] B. Found, D. Rogers, and C. Bird. 2012. Documentation of Forensic Handwriting Method: A Modular Approach. Technical Report. Victoria Police Forensic Services Department, Victoria, Australia.

[87] M. Frank, R. Biedert, E. Ma, I. Martinovic, and D. Song. 2013. Touchalytics: On the applicability of touchscreen input as a behavioral biometric for continuous authentication. IEEE Transactions on Information Forensics and Security 8, 1 (2013), 136–148.

[88] M. Freire, J. Fierrez, M. Martinez-Diaz, and J. Ortega-Garcia. 2007. On the applicability of off-line signatures to the fuzzy vault construction. In Int. Conf. on Document Analysis and Recognition, Vol. 2. IEEE, 1173–1177.

[89] C. Freitas, M. Morita, L. de Oliveira, E. Justino, A. Yacoubi, E. Lethelier, F. Bortolozzi, and R. Sabourin. 2000. Bases de dados de cheques bancários brasileiros. In XXVI Conferencia Latinoamericana de Informatica. 209–226.

[90] E. Frias-Martinez, A. Sanchez, and J. Velez. 2006. Support vector machines versus multi-layer perceptrons for efficient off-line signature recognition. Engineering Applications of Artificial Intelligence 19, 6 (2006), 693–704.

[91] J. Galbally, M. Diaz-Cabrera, M. A Ferrer, M. Gomez-Barrero, A. Morales, and

J. Fierrez. 2015. On-line signature recognition through the combination of real dynamic data and synthetically generated static data. Pattern Recognition 48, 9 (2015), 2921–2934.

[92] J. Galbally, J. Fierrez, M. Martinez-Diaz, and J. Ortega-Garcia. 2009. Improving the enrollment in dynamic signature verfication with synthetic samples. In 10th Int. Conf. on Document Analysis and Recognition (ICDAR'09). 1295–1299.

[93] J. Galbally, J. Fierrez, J. Ortega-Garcia, and R. Plamondon. 2012. Synthetic on-line signature generation. Part II: Experimental validation. Pattern Recognition 45 (2012), 2622–2632.

[94] J. Galbally, M. Gomez-Barrero, and A. Ross. 2017. Accuracy evaluation of handwritten signature verification: Rethinking the random-skilled forgeries dichotomy. In Int. Joint Conf. on Biometrics (IJCB'17). IEEE, 302–310.

[95] J. Galbally, M. Martinez, and J. Fierrez. 2013. Aging in biometrics: An experimental analysis on on-line signature. PLoS ONE 8, 7 (2013), e69897.

[96] J. Galbally, R. Plamondon, J. Fierrez, and J. Ortega-Garcia. 2012. Synthetic on-line signature generation. Part I: Methodology and algorithms. Pattern Recognition 45 (2012), 2610–2621.

[97] S. Garcia-Salicetti, N. Houmani, and B. Dorizzi. 2008. A client-entropy measure for on-line signatures. In Biometrics Symposium. IEEE, 83–88.

[98] S. Ghandali and M. E. Moghaddam. 2008. A method for off-line Persian signature identification and verification using DWT and image fusion. In Int. Symposium on Signal Processing and Information Technology. IEEE, 315–319.

[99] S. Ghandali and M. E. Moghaddam. 2009. Off-line Persian signature identification and verification based on image registration and fusion. Journal of Multimedia 4 (2009), 137–144. https://www.semanticscholar.org/paper/Off-LinePersian-Signature-Identification-and-Based-Ghandali-Moghaddam/7076a5b9b8e5bc121a3e8f57b236ce5651359988.

[100] M. Gomez-Barrero, J. Fierrez, and J. Galbally. 2016. Variable-length template protection based on homomorphic encryption with application to signature biometrics. In 4th Int. Workshop on Biometrics and Forensics. IEEE, 1–6.

[101] M. Gomez-Barrero, J. Galbally, J. Fierrez, and J. Ortega-Garcia. 2011. Hill-climbing attack based on the uphill simplex algorithm and its application to signature verification. In Proc. European Workshop on Biometrics and Identity Management (BioID'11) (LNCS-6583). Springer, 83–94.

[102] M. Gomez-Barrero, J. Galbally, J. Fierrez, J. Ortega-Garcia, and R. Plamondon. 2015. Enhanced on-line signature verification based on skilled forgery detection using sigma-lognormal features. In Conf. on Biometrics (ICB'15). 501–506. ACM Computing Surveys, Vol. 51, No. 6, Article 117. Publication date: January 2019. A Perspective Analysis of Handwritten Signature Technology 117:33

[103] C. Gruber, T. Gruber, S. Krinninger, and B. Sick. 2010. Online signature verification with support vector machines based on LCSS kernel functions. IEEE Transactions on Systems, Man, and Cybernetics Part B: Cybernetics 40, 4 (2010), 1088–1100.

[104] Y. Guerbai, Y. Chibani, and B. Hadjadji. 2015. The effective use of the one-class SVM classifier for handwritten signature verification based on writer-

independent parameters. Pattern Recognition 48, 1 (2015), 103–113.

[105] J. Guerra-Casanova, G. Sánchez-Ávila, C. Bailador, and A. de Santos-Sierra. 2011. Time series distances measures to analyze in-air signatures to authenticate users on mobile phones. In IEEE Int. Carnahan Conf. on Security Technology (ICCST'11). IEEE, 1–7.

[106] R. M. Guest, O. M. Hurtado, and O. Henniger. 2014. Assessment of methods for image recreation from signature time-series data. IET Biometrics 3, 3 (2014), 159–166.

[107] D. S. Guru, K. S. Manjunatha, S. Manjunath, and M. T. Somashekara. 2017. Interval valued symbolic representation of writer dependent features for online signature verification. Expert Systems with Applications 80 (2017), 232–243.

[108] D. S. Guru and H. N. Prakash. 2009. Online signature verification and recognition: An approach based on symbolic representation. IEEE Transactions on Pattern Analysis and Machine Intelligence 31, 6 (2009), 1059–1073.

[109] L. G. Hafemann, R. Sabourin, and L. S. Oliveira. 2016. Writer-independent feature learning for offline signature verification using deep convolutional neural networks. In Int. Joint Conf. on Neural Networks (IJCNN'16). IEEE, 2576–2583.

[110] L. G. Hafemann, R. Sabourin, and L. S. Oliveira. 2017. Learning features for offline handwritten signature verification using deep convolutional neural networks. Pattern Recognition 70 (2017), 163–176.

[111] L. G. Hafemann, R. Sabourin, and L. S. Oliveira. 2017. Offline handwritten signature verification - literature review. In 7th Int. Conf. on Image Processing Theory, Tools and Applications (IPTA'17). 1–5.

[112] A. Hamadene and Y. Chibani. 2016. One-class writer-independent offline signature verification using feature dissimilarity thresholding. IEEE Transactions on Information Forensics and Security 11, 6 (2016), 1226–1238.

[113] A. B. Hernandez, A. Fischer, and R. Plamondon. 2015. Omega-lognormal analysis of oscillatory movements as a function of brain stroke risk factors. In 17th Biennial Conf. of the Int. Graphonomics Society (IGS'15).

[114] C. Holz and P. Baudisch. 2013. Fiberio: A touchscreen that senses fingerprints. In Proc. 26th Annual ACM Symp. on User Interface Software and Technology. ACM, 41–50.

[115] N. Houmani, S. Garcia, B. Dorizzi, J. Montalvao, J. Canuto, M. Andrade, Y. Qiao, X. Wang, T. Scheidat, A. Makrushin, D. Muramatsu, J. Putz, M. Kudelski, M. Faundez, J. Pascual, V. Cardeñoso, C. Vivaracho, E. Argones Rúa, J. Alba, A. Kholmatov, and B. Yanikoglu. 2011. BioSecure signature evaluation campaign (ESRA'2011): Evaluating systems on quality-based categories of skilled forgeries. In 2011 Int. Joint Conf. on Biometrics (IJCB'11). 1–10.

[116] N. Houmani and S. Garcia-Salicetti. 2016. On hunting animals of the biometric menagerie for online signature. PLoS ONE 11, 4 (2016), e0151691.

[117] N. Houmani, S. Garcia-Salicetti, and B. Dorizzi. 2009. On assessing the robustness of pen coordinates, pen pressure and pen inclination to time variability with personal entropy. In IEEE 3rd Int. Conf. on Biometrics: Theory, Applications, and Systems. 1–6.

[118] N. Houmani, S. Garcia-Salicetti, B. Dorizzi, and M. El-Yacoubi. 2010. On-line

signature verification on a mobile platform. In Int. Conf. on Mobile Computing, Applications, and Services. Springer, 396–400.

[119] N. Houmani, A. Mayoue, S. Garcia-Salicetti, B. Dorizzi, M. I. Khalil, M. N. Moustafa, H. Abbas, D. Muramatsu, B. Yanikoglu, A. Kholmatov, M. Martinez-Diaz, J. Fierrez, J. Ortega-Garcia, J. Roure Alcobé, J. Fabregas, M. FaundezZanuy, J. M. Pascual-Gaspar, V. Cardeñoso-Payo, and C. Vivaracho-Pascual. 2012. BioSecure signature evaluation campaign (BSEC2009): Evaluating online signature algorithms depending on the quality of signatures. Pattern Recognition 45, 3 (2012), 993–1003.

[120] N. Houmani, S. Garcia Salicetti, and B. Dorizzi. 2008. A novel personal entropy measure confronted with online signature verification systems' performance. In 2nd IEEE Intl. Conf. on Biometrics: Theory, Applications and Systems (BTAS'08). 1–6.

[121] A. M. Huber, and R. A. Headrick. 1999. Handwriting Identification: Facts and Fundamentals. Boca Raton, CRC Press.

[122] M. T. Ibrahim, M. Kyan, M. A. Khan, and L. Guan. 2010. On-line signature verification using 1-D velocity-based directional analysis. In 20th International Conference on Pattern Recognition (ICPR'10). 3830–3833.

[123] D. Impedovo and G. Pirlo. 2008. Automatic signature verification: The state of the art. IEEE Transactions on Systems, Man, and Cybernetics 38, 5 (2008), 609–635.

[124] D. Impedovo and G. Pirlo. 2010. On-line signature verification by stroke-dependent representation domains. In 12th Int. Conf. on Frontiers in Handwriting Recognition (ICFHR'10). 623–627.

[125] D. Impedovo and G. Pirlo. 2018. Dynamic handwriting analysis for the assessment of neurodegenerative diseases: A pattern recognition perspective. IEEE Reviews in Biomedical Engineering (2018). DOI:10.1109/RBME.2018.2840679

[126] D. Impedovo, G. Pirlo, and R. Plamondon. 2012. Handwritten signature verification: New advancements and open issues. In Int. Conf. on Frontiers in Handwriting Recognition (ICFHR'12). 367–372. ACM Computing Surveys, Vol. 51, No. 6, Article 117. Publication date: January 2019. 117:34 M. Diaz et al.

[127] M. A. Ismail and S. Gad. 2000. Off-line arabic signature recognition and verification. Pattern Recognition 33, 10 (2000), 1727–1740.

[128] T. Ito, W. Ohyama, T. Wakabayashi, and F. Kimura. 2012. Combination of signature verification techniques by SVM. In 13th Int. Conf. on Frontiers in Handwriting Recognition (ICFHR'12). IEEE, 430–433.

[129] J. Ji, C. B. Chen, and X. Chen. 2010. Off-line Chinese signature verification: Using weighting factor on similarity computation. In 2nd Int. Conf. on E-Business and Information System Security. 1–4.

[130] J. Ji and X. Chen. 2009. Off-line Chinese signature verification segmentation and feature extraction. In Int. Conf. on Computational Intelligence and Software Engineering (CiSE'09). 1–4.

[131] J. Ji, Z. D. Lu, and X. Chen. 2009. Similarity computation based on feature extraction for off-line Chinese signature verification. In 6th Int. Conf. on Fuzzy Systems and Knowledge Discovery, Vol. 1. 291–295.

[132] A. Juels and M. Sudan. 2006. A fuzzy vault scheme. Designs, Codes and Cryptography 38, 2 (2006), 237–257.

[133] C. Kahindo, S. Garcia-Salicetti, and N. Houmani. 2015. A signature complexity measure to select reference signatures for online signature verification. In Int. Conf. of the Biometrics Special Interest Group (BIOSIG'15). IEEE, 1–8.

[134] M. K. Kalera, S. Srihari, and A. Xu. 2004. Offline signature verification and identification using distance statistics. International Journal of Pattern Recognition and Artificial Intelligence 18, 7 (2004), 1339–1360.

[135] N. S. Kamel, S. Sayeed, and G. A. Ellis. 2008. Glove-based approach to online signature verification. IEEE Transactions on Pattern Analysis and Machine Intelligence 30, 6 (2008), 1109–1113.

[136] Y. Kamihira, W. Ohyama, T. Wakabayashi, and F. Kimura. 2013. Improvement of Japanese signature verification by segmentation-verification. In 12th Int. Conf. on Document Analysis and Recognition (ICDAR'13). 379–382.

[137] Y. Kamihira, W. Ohyama, T. Wakabayashi, and F. Kimura. 2013. Improvement of Japanese signature verification by combined segmentation verification approach. In 2nd IAPR Asian Conf. on Pattern Recognition. 501–505.

[138] Y. Kawazoe, W. Ohyama, T. Wakabayashi, and F. Kimura. 2010. Improvement of on-line signature verification based on gradient features. In Int. Conf. on Frontiers in Handwriting Recognition (ICFHR'10). IEEE, 410–414.

[139] H. Khalajzadeh, M. Mansouri, and M. Teshnehlab. 2012. Persian signature verification using convolutional neural networks. International Journal of Engineering Research and Technology 1, 2 (2012), 7–12.

[140] A. Kholmatov and B. Yanikoglu. 2008. SUSIG: An on-line signature database, associated protocols and benchmark results. Pattern Analysis and Applications 12, 3 (2008), 227–236.

[141] H. Kim, S. Lee, and K. Yun. 2011. Capacitive tactile sensor array for touch screen application. Sensors and Actuators A: Physical 165, 1 (2011), 2–7.

[142] B. Kovari and H. Charaf. 2013. A study on the consistency and significance of local features in off-line signature verification. Pattern Recognition Letters 34, 3 (2013), 247–255.

[143] R. P. Krish, J. Fierrez, J. Galbally, and M. Martinez-Diaz. 2013. Dynamic signature verification on smart phones. In Int. Conf. on Practical Applications of Agents and Multi-Agent Systems (PAAMS'13). Springer, 213–222.

[144] R. Kumar, J. D. Sharma, and B. Chanda. 2012. Writer-independent off-line signature verification using surroundedness feature. Pattern Recognition Letters 33, 3 (2012), 301–308.

[145] S. Lai, L. Jin, and W. Yang. 2017. Online signature verification using recurrent neural network and length-normalized path signature. In 14th IAPR Int. Conf. on Document Analysis and Recognition (ICDAR'17). 1–6.

[146] A. Lanitis. 2009. A survey of the effects of aging on biometric identity verification. International Journal of Biometrics 2, 1 (2009), 34–52. [147] B. Li. 2016. The hot issues and future direction of forensic document examination in China. Journal of Forensic Science and Medicine 2, 1 (2016), 22–27.

[148] N. Li, J. Liu, Q. Li, X. Luo, and J. Duan. 2016. Online signature verification based on biometric features. In 49th Hawaii Int. Conf. on System Sciences (HICSS'16). 5527–5534. [149] Y. Liu, Z. Yang, and L. Yang. 2015. Online signature

verification based on DCT and sparse representation. IEEE Transactions on Cybernetics 45, 11 (2015), 2498–2511.

[150] M. Liwicki, M. Imran Malik, L. Alewijnse, E. van den Heuvel, and B. Found. 2012. ICFHR 2012 competition on automatic forensic signature verification (4NsigComp 2012). In 13th Int. Conf. on Frontiers in Handwriting Recognition (ICFHR'12). IEEE, 823–828.

[151] M. Liwicki, M. Imran Malik, C. E. van den Heuvel, X. Chen, C. Berger, R. Stoel, M. Blumenstein, and B. Found. 2011. Signature verification competition for online and offline skilled forgeries (SigComp2011). In 11th Int. Conf. on Document Analysis and Recognition (ICDAR'11). 1480–1484.

[152] M. Liwicki, C. E. van den Heuvel, B. Found, and M. I. Malik. 2010. Forensic signature verification competition 4NSigComp2010 - detection of simulated and disguised signatures. In 12th Int. Conf. on Frontiers in Handwriting Recognition (ICFHR'10). 715–720.

[153] M. Lorch. 2013. Written language production disorders: Historical and recent perspectives. Current Neurology and Neuroscience Reports 13, 8 (2013), 369. ACM Computing Surveys, Vol. 51, No. 6, Article 117. Publication date: January 2019. A Perspective Analysis of Handwritten Signature Technology 117:35

[154] O. M. Hurtado, R. Guest, and T. Chatzisterkotis. 2016. A new approach to automatic signature complexity assessment. In IEEE Int. Carnahan Conf. on Security Technology (ICCST'16). IEEE, 1–7.

[155] O. M. Hurtado, R. Guest, S. V. Stevenage, and G. J. Neil. 2014. The relationship between handwritten signature production and personality traits. In IEEE Int. Joint Conf. on Biometrics (IJCB'14). IEEE, 1–8.

[156] E. Maiorana and P. Campisi. 2010. Fuzzy commitment for function-based signature template protection. IEEE Signal Processing Letters 17, 3 (2010), 249–252.

[157] E. Maiorana, P. Campisi, J. Fierrez, J. Ortega-Garcia, and A. Neri. 2010. Cancelable templates for sequence-based biometrics with application to on-line signature recognition. IEEE Transactions on Systems, Man, and Cybernetics Part A: Systems and Humans 40, 3 (2010), 525–538.

[158] M. I. Malik, S. Ahmed, A. Dengel, and M. Liwicki. 2012. A signature verification framework for digital pen applications. In 10th IAPR International Workshop on Document Analysis Systems (DAS'12). IEEE, 419–423.

[159] M. I. Malik, S. Ahmed, M. Liwicki, and A. Dengel. 2013. Freak for real time forensic signature verification. In 12th Int. Conf. on Document Analysis and Recognition (ICDAR'13). IEEE, 971–975.

[160] M. I. Malik, S. Ahmed, A. Marcelli, U. Pal, M. Blumenstein, L. Alewijns, and M. Liwicki. 2015. ICDAR2015 competition on signature verification and writer identification for on- and off-line skilled forgeries (SigWIcomp2015). In 13th Int. Conf. on Document Analysis and Recognition (ICDAR'15). 1186–1190.

[161] M. I. Malik, M. Liwicki, L. Alewijnse, W. Ohyama, M. Blumenstein, and B. Found. 2013. Signature verification and writer identification competitions for on- and offline skilled forgeries (SigWiComp2013). In 12th Int. Conf. on Document Analysis and Recognition (ICDAR'13). 1477–1483.

[162] M. I. Malik, M. Liwicki, A. Dengel, and B. Found. 2014. Man vs. machine: A

comparative analysis for signature verification. Journal of Forensic Document Examination 24 (2014), 21–35.

[163] M. I. Malik, M. Liwicki, A. Dengel, S. Uchida, and V. Frinken. 2014. Automatic signature stability analysis and verification using local features. In 14th Int. Conf. on Frontiers in Handwriting Recognition (ICFHR'14). IEEE, 621–626.

[164] S. Mamoun. 2012. Off-line arabic signature verification using geometrical features. In National Workshop on Information Assurance Research. 1–6.

[165] K. S. Manjunatha, S. Manjunath, D. S. Guru, and M. T. Somashekara. 2016. Online signature verification based on writer dependent features and classifiers. Pattern Recognition Letters 80 (2016), 129–136.

[166] M. Martinez-Diaz, J. Fierrez, J. Galbally, and J. Ortega-Garcia. 2008. Towards mobile authentication using dynamic signature verification: Useful features and performance evaluation. In 19th Int. Conf. on Pattern Recognition (ICPR'08). IEEE, 1–5.

[167] M. Martinez-Diaz, J. Fierrez, R. P. Krish, and J. Galbally. 2014. Mobile signature verification: Feature robustness and performance comparison. IET Biometrics 3, 4 (2014), 267–277.

[168] H. R. Martinez-Hernandez and E. D. Louis. 2014. Macrographia in essential tremor: A study of patients with and without rest tremor. Movement Disorders 29, 7 (2014), 960–961.

[169] K. Matsuda, W. Ohyama, T. Wakabayashi, and F. Kimura. 2016. Effective random-impostor training for combined segmentation signature verification. In 15th Int. Conf. on Frontiers in Handwriting Recognition (ICFHR'16). IEEE, 489– 494.

[170] A. J. Mauceri. 1965. Feasibility studies of personal identification by signature verification. Technical Report SID65– 24. North American Aviation.

[171] A. Mendaza-Ormaza, O. M. Hurtado, R. Sánchez-Reillo, and J. Uriarte. 2010. Analysis on the resolution of the different signals in an on-line handwritten signature verification system applied to portable devices. In IEEE Int. Carnahan Conf. on Security Technology (ICCST'10). 341–350.

[172] A. Mendaza-Ormaza, O. Miguel-Hurtado, R. Blanco-Gonzalo, and F.-J. Diez-Jimeno. 2011. Analysis of handwritten signature performances using mobile devices. In IEEE Int. Carnahan Conf. on Security Technology (ICCST'11). 1–6.

[173] M. Meng, X. Xi, and Z. Luo. 2008. On-line signature verification based on support vector data description and genetic algorithm. In 7th World Congress on Intelligent Control and Automation. 3778–3782.

[174] Y. Meng, D. S. Wong, R. Schlegel, and L. Kwok. 2012. Touch gestures based biometric authentication scheme for touchscreen mobile phones. In Int. Conf. on Information Security and Cryptology. Springer, 331–350.

[175] L. A. Mohammed, B. Found, M. Caligiuri, and D. Rogers. 2011. The dynamic character of disguise behavior for textbased, mixed, and stylized signatures. Journal of Forensic Sciences 56, s1 (2011), S136–S141. https://onlinelibrary. wiley.com/doi/pdf/10.1111/j.1556-4029.2010.01584. x.

[176] A. Morales, D. Morocho, J. Fierrez, and R. Vera-Rodriguez. 2017. Signature authentication based on human intervention: Performance and complementarity with automatic systems. IET Biometrics 6, 4 (2017), 307–315. [177] S. Nam, C. Seo, and

D. Choi. 2016. Mobile finger signature verification robust to skilled forgery. Journal of the Korea Institute of Information Security & Cryptology 26, 5 (2016), 1161–1170.

[178] L. Nanni, E. Maiorana, A. Lumini, and P. Campisi. 2010. Combining local, regional and global matchers for a template protected on-line signature verification system. Expert Systems with Applications 37, 5 (2010), 3676–3684. ACM Computing Surveys, Vol. 51, No. 6, Article 117. Publication date: January 2019. 117:36 M. Diaz et al.

[179] T. V. Nguyen, N. Sae-Bae, and N. Memon. 2017. DRAW-A-PIN: Authentication using finger-drawn PIN on touch devices. Computers & Security 66 (2017), 115–128.

[180] V. Nguyen and M. Blumenstein. 2010. Techniques for static handwriting trajectory recovery: A survey. In Proc. of the 9th IAPR Int. Workshop on Document Analysis Systems (DAS'10). 463–470.

[181] V. Nguyen, M. Blumenstein, and G. Leedham. 2009. Global features for the off-line signature verification problem. In 10th Int. Conf. on Document Analysis and Recognition (ICDAR'09). 1300–1304.

[182] A. Nordgaard and B. Rasmusson. 2012. The likelihood ratio as value of evidence - more than a question of numbers. Law, Probability and Risk 11, 4 (2012), 303–315.

[183] W. Ohyama, Y. Ogi, T. Wakabayashi, and F. Kimura. 2015. Multilingual signature-verification by generalized combined segmentation verification. In 13th Int. Conference on Document Analysis and Recognition (ICDAR'15). 811–815.

[184] M. Okawa. 2017. KAZE features via Fisher vector encoding for offline signature verification. In Int. Joint Conf. on Biometrics (IJCB'17). IEEE, 10–15.

[185] E. Onofri, M. Mercuri, M. Salesi, S. Ferrara, G. M. Troili, C. Simeone, M. R. Ricciardi, S. Ricci, and T. Archer. 2013. Dysgraphia in relation to cognitive performance in patients with Alzheimer's disease. Journal of Intellectual DisabilityDiagnosis and Treatment 1, 2 (2013), 113–124.

[186] C. O'Reilly and R. Plamondon. 2009. Development of a sigma-lognormal representation for on-line signatures. Pattern Recognition 42, 12 (2009), 3324–3337.

[187] C. O'Reilly and R. Plamondon. 2011. Impact of the principal stroke risk factors on human movements. Human Movement Science 30, 4 (2011), 792–806.

[188] C. O'Reilly and R. Plamondon. 2013. Agonistic and antagonistic interaction in speed/accuracy tradeoff: A deltalognormal perspective. Human Movement Science 32, 5 (2013), 1040–1055.

[189] C. O'Reilly, R. Plamondon, and L-H. Lebrun. 2014. Linking brain stroke risk factors to human movement features for the development of preventive tools. Frontiers in Aging Neuroscience 6, 150 (2014), 1–13.

[190] J. Ortega-Garcia, J. Fierrez, F. Alonso-Fernandez, J. Galbally, M. Freire, J. Gonzalez-Rodriguez, C. Garcia-Mateo, J.- L. Alba-Castro, E. Gonzalez-Agulla, E. Otero-Muras, S. Garcia-Salicetti, L. Allano, B. Ly-Van, B. Dorizzi, J. Kittler, T. Bourlai, N. Poh, F. Deravi, M. Ng, M. Fairhurst, J. Hennebert, A. Humm, M. Tistarelli, L. Brodo, J. Richiardi, A. Drygajlo, H. Ganster, F. M. Sukno, S.-K. Pavani, A. Frangi, L. Akarun, and A. Savran. 2010. The multi-scenario multienvironment

BioSecure multimodal database (BMDB). IEEE Transactions on Pattern Analysis and Machine Intelligence 32, 6 (2010), 1097–1111.

[191] J. Ortega-Garcia, J. Fierrez-Aguilar, D. Simon, J. Gonzalez, M. Faundez-Zanuy, V. Espinosa, A. Satue, I. Hernaez, J.-J. Igarza, C. Vivaracho, D. Escudero, and Q. I. Moro. 2003. MCYT baseline corpus: A bimodal biometric database. IEE Proceedings-Vision, Image and Signal Processing 150, 6 (2003), 395–401.

[192] S. Pal, A. Alaei, U. Pal, and M. Blumenstein. 2016. Performance of an off-line signature verification method based on texture features on a large indicscript signature dataset. In 12th IAPR Workshop on Document Analysis Systems (DAS'16). IEEE, 72–77.

[193] S. Pal, A. Reza, U. Pal, and M. Blumenstein. 2013. SVM and NN based offline signature verification. International Journal of Computational Intelligence and Applications 12, 4 (2013), 1340004.

[194] S. Pal, A. Alaei, U. Pal, and M. Blumenstein. 2011. Off-line signature identification using background and foreground information. In Int. Conf. on Digital Image Computing Techniques and Applications (DICTA'11). IEEE, 672–677.

[195] S. Pal, A. Alaei, U. Pal, and M. Blumenstein. 2012. Multi-script off-line signature identification. In Int. Conf. on Hybrid Intelligent Systems (HIS'12). 236–240.

[196] S. Pal, M. Blumenstein, and U. Pal. 2011. Non-English and non-Latin signature verification systems a survey. In 1st Int. Workshop on Automated Forensic Handwriting Analysis (AFHA'11). 1–5.

[197] S. Pal, U. Pal, and M. Blumenstein. 2012. Off-line English and Chinese signature identification using foreground and background features. In Int. Joint Conf. on Neural Networks (IJCNN'12). 1–7.

[198] S. Pal, U. Pal, and M. Blumenstein. 2013. Hindi and English off-line signature identification and verification. In Proc. of Int. Conf. on Advances in Computing. Springer, New Delhi, 905–910.

[199] S. Pal, U. Pal, and M. Blumenstein. 2013. Multi-script off-line signature verification: A two stage approach. In 2nd Int. Workshop on Automated Forensic Handwriting Analysis (AFHA'13). 31–35.

[200] S. Pal, U. Pal, and M. Blumenstein. 2013. Off-line verification technique for Hindi signatures. IET Biometrics 2, 4 (2013), 182–190.

[201] M. Parodi and J. C. Gómez. 2014. Legendre polynomials based feature extraction for online signature verification. Consistency analysis of feature combinations. Pattern Recognition 47, 1 (2014), 128–140.

[202] M. Parodi, J. C. Gómez, and L. Alewijnse. 2014. Automatic online signature verification based only on FHE features: An oxymoron? In 14th Int. Conf. on Frontiers in Handwriting Recognition (ICFHR'14). 73–78.

[203] M. Parodi, J. C. Gómez, M. Liwicki, and L. Alewijnse. 2013. Orthogonal function representation for online signature verification: Which features should be looked at? IET Biometrics 2, 4 (2013), 137–150. ACM Computing Surveys, Vol. 51, No. 6, Article 117. Publication date: January 2019. A Perspective Analysis of Handwritten Signature Technology 117:37

[204] A. Parziale, S. G. Fuschetto, and A. Marcelli. 2013. Exploiting stability

regions for online signature verification. In New Trends in Image Analysis and Processing. Lecture Notes in Computer Science, Vol. 8158. Springer, 112–121.

[205] M. Piekarczyk and M. R. Ogiela. 2013. Matrix-based hierarchical graph matching in off-line handwritten signatures recognition. In 2nd IAPR Asian Conf. on Pattern Recognition. IEEE, 897–901.

[206] G. Pirlo, V. Cuccovillo, M. Diaz-Cabrera, D. Impedovo, and P. Mignone. 2015. Multidomain verification of dynamic signatures using local stability analysis. IEEE Transactions on Human-Machine Systems 45, 6 (2015), 805–810. https://ieeexplore.ieee.org/document/7145458.

[207] G. Pirlo, M. Diaz, M. A. Ferrer, D. Impedovo, F. Occhionero, and U. Zurlo. 2015. Early diagnosis of neurodegenerative diseases by handwritten signature analysis. In 1st Workshop on Image-based Smart City Applications: Int. Conf. on Image Analysis and Processing (ICIAP'15). 290–297.

[208] G. Pirlo and D. Impedovo. 2013. Cosine similarity for analysis and verification of static signatures. IET Biometrics 2, 4 (2013), 151–158.

[209] G. Pirlo and D. Impedovo. 2013. Verification of static signatures by optical flow analysis. IEEE Transactions on HumanMachine Systems 43, 5 (2013), 499–505.

[210] G. Pirlo, D. Impedovo, R. Plamondon, C. O'Reilly, A. Cozzolongo, R. Gravinese, and A. Rollo. 2013. Stability of dynamic signatures: From the representation to the generation domain. In New Trends in Image Analysis and Processing. Lecture Notes in Computer Science, Vol. 8158. Springer, 122–130.

[211] G. Pirlo, D. Impedovo, and T. Ferranti. 2015. Stability/complexity analysis of dynamic handwritten signatures. In 17th Biennial Conf. of the Int. Graphonomics Society (IGS'15).

[212] G. Pirlo, D. Impedovo, R. Plamondon, and C. O'Reilly. 2014. Stability analysis of online signatures in the generation domain. In Advances in Digital Handwritten Signature Processing. A Human Artefact for e-Society. 1–12.

[213] R. Plamondon, C. O'Reilly, C. Rémi, and T. Duval. 2013. The lognormal handwriter: Learning, performing, and declining. Frontiers in Psychology 4 (2013), 945.

[214] R. Plamondon and G. Lorette. 1989. Automatic signature verification and writer identification - the state of the art. Pattern Recognition 22, 2 (1989), 107–131.

[215] R. Plamondon, C. O'Reilly, and C. Ouellet-Plamondon. 2014. Strokes against stroke - strokes for strides. Pattern Recognition 47, 3 (2014), 929–944.

[216] R. Plamondon, G. Pirlo, E. Anquetil, C. Remi, H-L. Teulings, and M. Nakagawa. 2018. Personal digital bodyguards for e-Security, e-Learning and e-Health: A prospective survey. Pattern Recognition 81, 2018 (2018), 633–659. https://doi.org/10.1016/j.patcog.2018.04.012.

[217] R. Plamondon, G. Pirlo, and D. Impedovo. 2014. Online signature verification. In Handbook of Document Image Processing and Recognition. Springer, 917–947.

[218] R. Plamondon and S. N. Srihari. 2000. On-line and off-line handwriting recognition: A comprehensive survey. IEEE Transactions on Pattern Analysis and Machine Intelligence 22, 1 (2000), 63–84.

[219] C. Rabasse, R. M. Guest, and M. C. Fairhurst. 2008. A new method for the synthesis of signature data with natural variability. IEEE Transactions on Systems,

Man, and Cybernetics, Part B: Cybernetics 38, 3 (2008), 691–699.

[220] H. Rantzsch, H. Yang, and C. Meinel. 2016. Signature embedding: Writer independent offline signature verification with deep metric learning. In Int. Symposium on Visual Computing. Springer, 616–625.

[221] C. Rathgeb and A. Uhl. 2011. A survey on biometric cryptosystems and cancelable biometrics. EURASIP Journal on Information Security 2011, 1 (2011), 3.

[222] M. Renier, F. Gnoato, A. Tessari, M. Formilan, F. Busonera, P. Albanese, G. Sartori, and A. Cester. 2016. A correlational study between signature, writing abilities and decision-making capacity among people with initial cognitive impairment. Aging Clinical and Experimental Research 28, 3 (2016), 505–511.

[223] D. Rivard, E. Granger, and R. Sabourin. 2013. Multifeature extraction and selection in writer-independent off-line signature verification. International Journal on Document Analysis and Recognition 16 (2013), 83–103.

[224] S. Rosenblum, M. Samuel, S. Zlotnik, I. Erikh, and I. Schlesinger. 2013. Handwriting as an objective tool for Parkinson's disease diagnosis. Journal of Neurology 260, 9 (2013), 2357–2361.

[225] S. Rosenblum, M. Samuel, S. Zlotnik, and I. Schlesinger. 2009. Handwriting performance measures of "real life" tasks: A comparison between the performance of patients with Parkinson's disease and controls. In 14th Conf. of the Int. Graphonomics Society (IGS'09). 43–47.

[226] S. Rosenblum, H. A. B. Simhon, and E. Gal. 2016. Unique handwriting performance characteristics of children with high-functioning autism spectrum disorder. Research in Autism Spectrum Disorders 23 (2016), 235–244.

[227] N. Sae-Bae and N. Memon. 2014. Online signature verification on mobile devices. IEEE Transactions on Information Forensics and Security 9, 6 (2014), 933–947.

[228] H. Saevanee and P. Bhatarakosol. 2008. User authentication using combination of behavioral biometrics over the touchpad acting like touch screen of mobile device. In Int. Conf. on Computer and Electrical Engineering (ICCEE'08). IEEE, 82–86.

ACM Computing Surveys, Vol. 51, No. 6, Article 117. Publication date: January 2019. 117:38 M. Diaz et al.

[229] B. Schafer and S. Viriri. 2009. An off-line signature verification system. In Int. Conf. on Signal and Image Processing Applications (ICSIPA'09). 95–100.

[230] M. Schulte-Austum. 2013. D-Scribe. In Presentation at Measurement Science and Standards in Forensic Handwriting Analysis Conf.

[231] R. Senatore and A. Marcelli. 2017. Do handwriting difficulties of Parkinson's patients depend on their impaired ability to retain the motor plan? A pilot studies. In 18th Biennial Conf. of the Int. Graphonomics Society (IGS'17). 139–142.

[232] Y. Serdouk, H. Nemmour, and Y. Chibani. 2017. Handwritten signature verification using the quad-tree histogram of templates and a support vector-based artificial immune classification. Image and Vision Computing 66 (2017), 26–35.

[233] E. Sesa-Nogueras, M. Faundez-Zanuy, and J. Mekyska. 2012. An information analysis of in-air and on-surface trajectories in online handwriting. Cognitive Computation 4, 2 (2012), 195–205.

[234] A. Sharma and S. Sundaram. 2016. An enhanced contextual DTW based

system for online signature verification using vector quantization. Pattern Recognition Letters 84 (2016), 22–28.

[235] A. Sharma and S. Sundaram. 2017. A novel online signature verification system based on GMM features in a DTW framework. IEEE Transactions on Information Forensics and Security 12, 3 (2017), 705–718.

[236] A. Sharma and S. Sundaram. 2018. On the exploration of information from the DTW cost matrix for online signature verification. IEEE Transactions on Cybernetics 48, 2 (2018), 611–624.

[237] O. Sinanović and Z. Mrkonjić. 2013. Poststroke writing and reading disorders. Sanamed 8, 1 (2013), 55–63.

[238] J. A. Small and N. Sandhu. 2008. Episodic and semantic memory influences on picture naming in Alzheimer's disease. Brain and Language 104, 1 (2008), 1–9.

[239] E. J. Smits, A. J. Tolonen, L. Cluitmans, M. van Gils, B. A. Conway, R. C. Zietsma, K. L. Leenders, and N. M. Maurits. 2014. Standardized handwriting to assess bradykinesia, micrographia and tremor in Parkinson's disease. PloS ONE 9, 5 (2014), e97614.

[240] A. Soleimani, B. N. Araabi, and K. Fouladi. 2016. Deep multitask metric learning for offline signature verification. Pattern Recognition Letters 80 (2016), 84–90.

[241] A. Soleimani, K. Fouladi, and B. N. Araabi. 2016. UTSig: A Persian offline signature dataset. IET Biometrics 6, 1 (2016), 1–8.

[242] M. Song and Z. Sun. 2014. An immune clonal selection algorithm for synthetic signature generation. Mathematical Problems in Engineering 2014 (2014), 1–12.

[243] S. Srihari. 2013. iFOX. In Presentation at Measurement Science and Standards in Forensic Handwriting Analysis Conf.

[244] C. D. Stefano, C. De Stefano, F. Fontanella, D. Impedovo, G. Pirlo, and A. Scotto di Freca. 2018. Handwriting analysis to support neurodegenerative diseases diagnosis: A review. Pattern Recognition Letters (2018).

[245] T. Steinherz, D. Doermann, E. Rivlin, and N. Intrator. 2009. Offline loop investigation for handwriting analysis. IEEE Transactions on Pattern Analysis and Machine Intelligence 31, 2 (2009), 193–209.

[246] R. Tolosana, R. Vera-Rodriguez, J. Fierrez, A. Morales, and J. Ortega-Garcia. 2017. Benchmarking desktop and mobile handwriting across COTS devices: The e-BioSign biometric database. PloS ONE 12, 5 (2017), e0176792.

[247] R. Tolosana, R. Vera-Rodriguez, J. Fierrez, and J. Ortega-Garcia. 2018. Exploring recurrent neural networks for online handwritten signature biometrics. IEEE Access (2018), 5128–5138.

[248] R. Tolosana, R. Vera-Rodriguez, R. Guest, J. Fierrez, and J. Ortega-Garcia. 2017. Complexity-based biometric signature verification. In Proc. 14th IAPR Int. Conference on Document Analysis and Recognition (ICDAR'17).

[249] R. Tolosana, R. Vera-Rodriguez, J. Ortega-Garcia, and J. Fierrez. 2015. Increasing the robustness of biometric templates for dynamic signature biometric systems. In Int. Carnahan Conf. on Security Technology (ICCST'15). IEEE, 229–234.

[250] R. Tolosana, R. Vera-Rodriguez, J. Ortega-Garcia, and J. Fierrez. 2015.

Preprocessing and feature selection for improved sensor interoperability in online biometric signature verification. IEEE Access 3 (2015), 478–489.

[251] R. Tolosana, R. Vera-Rodriguez, J. Ortega-Garcia, and J. Fierrez. 2015. Update strategies for HMM-based dynamic signature biometric systems. In Int. Workshop on Information Forensics and Security (WIFS'15). IEEE, 1–6.

[252] L. Y. Tseng and T. H. Huang. 1992. An online Chinese signature verification scheme based on the ART1 neural network. In Proc. 1992 IJCNN Int. Joint Conf. on Neural Networks (IJCNN'92), Vol. 3. 624–630.

[253] U. Uludag, S. Pankanti, S. Prabhakar, and A. Jain. 2004. Biometric cryptosystems: Issues and challenges. Proceedings of the IEEE 92, 6 (2004), 948–960.

[254] O. Ureche and R. Plamondon. 1999. Document transport, transfer and exchange: Security and commercial aspects. In 5th Int. Conf. on Document Analysis and Recognition (ICDAR'99). 585–588.

[255] J. F. Vargas, M. A. Ferrer, C. M. Travieso, and J. B. Alonso. 2011. Off-line signature verification based on grey level information using texture features. Pattern Recognition 44, 2 (2011), 375–385.

[256] F. Vargas, M. A. Ferrer, C. Travieso, and J. Alonso. 2007. Off-line handwritten signature GPDS-960 corpus. In Int. Conf. on Document Analysis and Recogn. (ICDAR'07), Vol. 2. 764–768. ACM Computing Surveys, Vol. 51, No. 6, Article 117. Publication date: January 2019. A Perspective Analysis of Handwritten Signature Technology 117:39

[257] R. J. Verduijn, E. van den Heuvel, and R. Stoel. 2011. Forensic requirements for automated handwriting analysis systems. In Biennial Conf. of the Int. Graphonomics Society (IGS'11). 132–135.

[258] C. Vielhauer. 2005. A behavioral biometrics. Public Service Review: European Union 9 (2005), 113–115.

[259] M. Vijayanunni. 1998. Planning for 2001 census of India based on 1991 census. In 18th Population Census Conf..

[260] M. Walch and D. Gantz. 2013. The forensic language-independent analysis system for handwriting identification (FLASH ID). Presentation at the Measurement Science and Standards In Forensic Handwriting Analysis Conference at The National Institute of Standards & Technology (NIST), Gaithersburg, Maryland.

[261] D. Wang, Y. Zhang, C. Yao, J. Wu, H. Jiao, and M. Liu. 2010. Toward force-based signature verification: A pen-type sensor and preliminary validation. IEEE Transactions on Instrumentation and Measurement 59, 4 (2010), 752–762.

[262] J.-S. Wang, Y.-L. Hsu, and J.-N. Liu. 2010. An inertial-measurement-unit-based pen with a trajectory reconstruction algorithm and its applications. IEEE Transactions on Industrial Electronics 57, 10 (2010), 3508–3521.

[263] K. Wang, Y. Wang, and Z. Zhang. 2011. On-line signature verification using graph representation. In 6th Int. Conf. on Image and Graphics (ICIG'11). 943–948.

[264] C. F. Whitelock, H. N. Agyepong, K. Patterson, and K. D. Ersche. 2015. Signing below the dotted line: Signature position as a marker of vulnerability for visuospatial processing difficulties. Neurocase 21, 1 (2015), 67–72.

[265] A. Woch and R. Plamondon. 2010. Characterization of bi-directional

movement primitives and their agonist antagonist synergy with the delta-lognormal model. Motor Control 14, 1 (2010), 1–25.

[266] A. Woch, R. Plamondon, and C. O'Reilly. 2011. Kinematic characteristics of bidirectional delta-lognormal primitives in young and older subjects. Human Movement Science 30, 1 (2011), 1–17.

[267] M. E. Yahyatabar, Y. Baleghi, and M. R. Karami. 2013. Online signature verification: A Persian-language specific approach. In 21st Iranian Conf. on Electrical Engineering (ICEE'13). 1–6.

[268] M. E. Yahyatabar and J. Ghasemi. 2017. Online signature verification using double-stage feature extraction modelled by dynamic feature stability experiment. IET Biometrics 6, 6 (2017), 393–401.

[269] J. H. Yan, S. Rountree, P. Massman, R. S. Doody, and H. Li. 2008. Alzheimer's disease and mild cognitive impairment deteriorate fine movement control. Journal of Psychiatric Research 42, 14 (2008), 1203–1212.

[270] K. Yasuda, B. Beckmann, and T. Nakamura. 2000. Brain processing of proper names. Aphasiology 14, 11 (2000), 1067–1089.

[271] D.-Y. Yeung, H. Chang, Y. Xiong, S. George, R. Kashi, T. Matsumoto, and G. Rigoll. 2004. SVC2004: First International signature verification competition. In Biometric Authentication. LNCS, Vol. 3072. Springer, 16–22.

[272] M. B. Yılmaz and B. Yanıkoğlu. 2016. Score level fusion of classifiers in off-line signature verification. Information Fusion 32 (2016), 109–119.

[273] F. J. Zareen and S. Jabin. 2016. Authentic mobile-biometric signature verification system. IET Biometrics 5, 1 (2016), 13–19.

[274] S. Zhang and F. Li. 2012. Off-line handwritten Chinese signature verification based on support vector machine multiple classifiers. In Advances in Electric and Electronics. Springer, 563–568.

[275] Z. Zhang, X. Liu, and Y. Cui. 2016. Multiphase offline signature verification system using deep convolutional generative adversarial networks. In 9th International Symposium on Computational Intelligence and Design (ISCID'16), Vol. 2. 103–107.

[276] N. Zheng, K. Bai, H. Huang, and H. Wang. 2014. You are how you touch: User verification on smartphones via tapping behaviors. In Int. Conf. on Network Protocols (ICNP'14). IEEE, 221–232.

[277] M. Zoghi and V. Abolghasemi. 2009. Persian signature verification using improved dynamic time warping-based segmentation and multivariate autoregressive modeling. In 15th Workshop on Statistical Signal Processing (SSP'09). 329–332.

[278] E. N. Zois, L. Alewijnse, and G. Economou. 2016. Offline signature verification and quality characterization using Poset-oriented grid features. Pattern Recognition 54 (2016), 162–177.